\documentclass[times,onecolumn,preprint,authoryear]{elsarticle} 
\graphicspath{{../images/}}
\usepackage{ycviu}
\usepackage{framed,multirow}

\usepackage{amssymb}
\usepackage{latexsym}

\usepackage{url}
\usepackage{xcolor}
\definecolor{newcolor}{rgb}{.8,.349,.1}
\usepackage{siunitx}
\usepackage{amsthm}
\usepackage{tikz}
\usepackage{graphicx}
\usetikzlibrary{positioning}

\usepackage[textwidth=2.8cm, textsize=scriptsize]{todonotes}
\usepackage{xspace}
\usepackage[caption=false,font=footnotesize]{subfig}
\usepackage{amsmath}
\usepackage[utf8]{inputenc}
\usepackage{cite}
\usepackage{algorithm}
\usepackage{algorithmic}
\usepackage{mathrsfs}

\usepackage{geometry}
\newcommand{\V}[1]{{\boldsymbol{#1}}}
\newcommand{\Vh}[1]{{\boldsymbol{\hat{#1}}}}

\newcommand{\Vu}[1]{{\boldsymbol{\underline{#1}}}}
\newcommand{\Vhu}[1]{{\boldsymbol{\hat{\underline{#1}}}}}

\definecolor{my_green}{RGB}{72, 178, 58}

\newcommand{\reva}[1]{\textcolor{black}{#1}}
\newcommand{\revb}[1]{\textcolor{black}{#1}}
\newcommand{\revc}[1]{\textcolor{black}{#1}}

\journal{Computer Vision and Image Understanding}

\begin{document}

\thispagestyle{empty}

\clearpage
\thispagestyle{empty}
\ifpreprint
  \vspace*{-1pc}
\fi

\ifpreprint
  \setcounter{page}{1}
\else
  \setcounter{page}{1}
\fi

\begin{frontmatter}

\title{Urban Surface Reconstruction in SAR Tomography by Graph-Cuts}

\author[1]{Cl\'ement \snm{Rambour}\corref{cor1}} 
\cortext[cor1]{Corresponding author: }
\ead{clement.rambour@telecom-paristech.fr}
\author[2]{Lo\"ic \snm{Denis}}
\author[1]{Florence \snm{Tupin}}
\author[3]{H\'el\`ene \snm{Oriot}}
\author[4]{Yue \snm{Huang}}
\author[4]{Laurent \snm{Ferro-Famil}}
\author[]{}

\address[1]{LTCI, T\'el\'ecom ParisTech,
	Universit\'e
	Paris-Saclay, 75013 Paris, France}
\address[2]{Univ Lyon, UJM-Saint-Etienne, CNRS,
	Institut
	d Optique Graduate School, Laboratoire Hubert Curien UMR 5516,
	F-42023,
	SAINT-ETIENNE, France}
\address[3]{ONERA, The French Aerospace
	Laboratory, 91761 Palaiseau, France}
\address[4]{Institute of Electronics and Telecommunications of Rennes (IETR), University of Rennes 1, Rennes, France}

\received{1 May 2013}
\finalform{10 May 2013}
\accepted{13 May 2013}
\availableonline{15 May 2013}
\communicated{S. Sarkar}

\begin{abstract}
	SAR (Synthetic Aperture Radar) tomography reconstructs 3-D volumes from stacks of SAR images.
	High resolution satellites such as TerraSAR-X provide images that can
	be combined to produce 3-D models. In urban areas, sparsity priors are
	generally enforced during the tomographic inversion process in order
	to retrieve the location of scatterers seen within a given radar
	resolution cell.
	However, such priors often miss parts of the urban surfaces.
	Those missing parts are typically regions of flat areas such as ground
	or rooftops.
	
	This paper introduces a surface segmentation algorithm
	based on the computation of the optimal cut in a flow network. This segmentation process can be included within the 3-D reconstruction framework in order to improve the recovery of urban surfaces.
	Illustrations on a TerraSAR-X tomographic dataset demonstrate the potential of the approach to produce a 3-D model of urban surfaces such as ground, façades and rooftops.
\end{abstract}


\end{frontmatter}


\section{Introduction}

SAR tomography is a remote sensing technique that can retrieve
3-D representations of diffuse environments such as forested or ice
areas \citep{Reigber}. 
Under the assumption that those
media are mostly homogeneous, the covariance matrix at each radar resolution cell
can satisfactorily be estimated by local averaging. Efficient spectral analysis
techniques can then be used to invert these covariance matrices and identify the height distribution of the radar reflectivity.

With the improvement of the available spatial resolution and the large time-series, SAR tomography is also being used to analyze complex
environments such as urban areas. Well-established estimators from the spectral analysis theory such as Capon beamforming \citep{capon} \citep{stoica},
MUSIC \citep{MUSIC},  Weighted Subspace Fitting (WSF) \citep{Viberg} or the more
recent SPICE \citep{spice} can be used to retrieve the 3-D distribution of the
backscattering targets from the radar covariance matrices. 
However, estimating the covariance matrices is very difficult in dense
urban configurations due to the spatial heterogeneity. Spatial averaging leads to blurring phenomena that bias the 3-D inversions.
Considering compressed sensing techniques,  the well-known approaches such as in  \citep{Zhu_CS_intro,Schirinzi_CS_intro} are based on $\ell_1$-norm minimization in the single-snapshot case, so they do not require to estimate the data covariance matrix. They nonetheless 
achieve
super-resolved estimation of the scatterers heights.
A recent \revc{extension of the conventional single look Compressed Sensing (CS)} has been proposed \citep{3Dinv} in order to also consider the 3-D relationship between scatterers through spatial regularizations.

With spaceborne sensors, these
techniques can be used to obtain a 3-D representation of vast areas on
the ground. Those 3-D models have numerous applications
such as town planning, city management or crisis monitoring. However, because of the large contrasts between back-scattered powers in urban areas (ground level and roof tops return much weaker echoes than structures on façades), the obtained 3-D models are accurate on some building walls but suffer from holes or large errors elsewhere.

Different techniques exist to recover the geometrical shapes of urban areas from tomographic reconstructions.  The proposed approaches generally use discretized
representations of the reconstructed volume to find a surface that represents
best the observed scene. Works \citep{3D_Zhu} and \citep{3D_Zhu_2} propose to extract
building shapes using numerous image processing techniques on the obtained
point cloud (region growing, directional filtering, clustering, polygon
fitting). In
 \citep{odhondt_andreas}, Ley {\it et al.} propose to minimize the Total Variation (TV) of the
height map that best fits the observed points. They introduce the
idea that along a given ray incoming from the sensor, only a single scatterer is expected to
be recovered. This hypothesis is particularly well-suited to urban areas where surfaces scatter most of the incoming wave so that down-stream surfaces are occluded (in the shadow). The observed scene can then be
described as a piecewise smooth surface. When minimizing the TV of the estimated
height map, holes in the structures are filled thanks to the spatial regularization constraint (by constant height areas). 

\reva{
Given the numerous successes of learning-based techniques in computer vision, one may wonder whether such techniques could be applied to the problem of reconstructing urban surfaces from tomographic SAR stacks.
SAR tomography consists of retrieving a 3-D image from a stack of 2-D measurements. It is fundamentally an inverse problem since it involves unmixing the signals from several scatterers that project within the same SAR resolution cell. It is important to include geometrical information that relates pixels in the 3-D reconstruction to resolution cells in the 2-D SAR images and to account for the delays (i.e., phase shifts) due to the acquisition geometry. Imposing geometrical priors to improve the 3-D reconstruction could be done using learned priors or a deep neural network, however, a strong interaction between the geometrical modeling corresponding to the forward model and the spatial regularization is needed. If only a segmentation approach is considered, starting from a given tomographic reconstruction, a difficulty arises from the reconstruction artifacts (sidelobes) that are very specific to the acquisition geometry and that might require retraining a learning-based segmentation technique for each tomographic stack to efficiently remove these specific artifacts. Given the difficulty in producing tomographic SAR stacks and associated 3D ground truths, or of simulating realistic SAR stacks from a 3D model (with the temporal and geometrical decorrelation effects), learning-based models do not seem to be, to date, a viable option. 
}

In this paper, we describe the urban scene as piecewise smooth surfaces and seek a segmentation of the urban surfaces as an optimal cut in a particular graph. To better express geometrical regularity properties, we segment the scene in
ground geometry. The so-called
ground geometry is composed of two axes that define the horizontal plane and
the vertical direction \textit{cf.} Fig. \ref{fig:tomo_geo}. We also use the
estimated 3-D reflectivity as input for the surface segmentation instead of providing a list of extracted 3-D points. As the proposed segmentation algorithm is
very generic, any SAR tomographic algorithm can be used as input.

Given the superior results obtained when combining the 3-D inversion algorithm \citep{3Dinv} with the surface segmentation, we propose to combine them both 
by	alternating a step of reflectivity estimation with a step of surface segmentation. We call this algorithm REDRESS for Alte{\bf R}nat{\bf E}d 3-{\bf D} {\bf
	RE}construction and {\bf S}urface
{\bf S}egmentation.

\begin{figure*}[!ht]
	\centering
	\includegraphics[width=\linewidth]{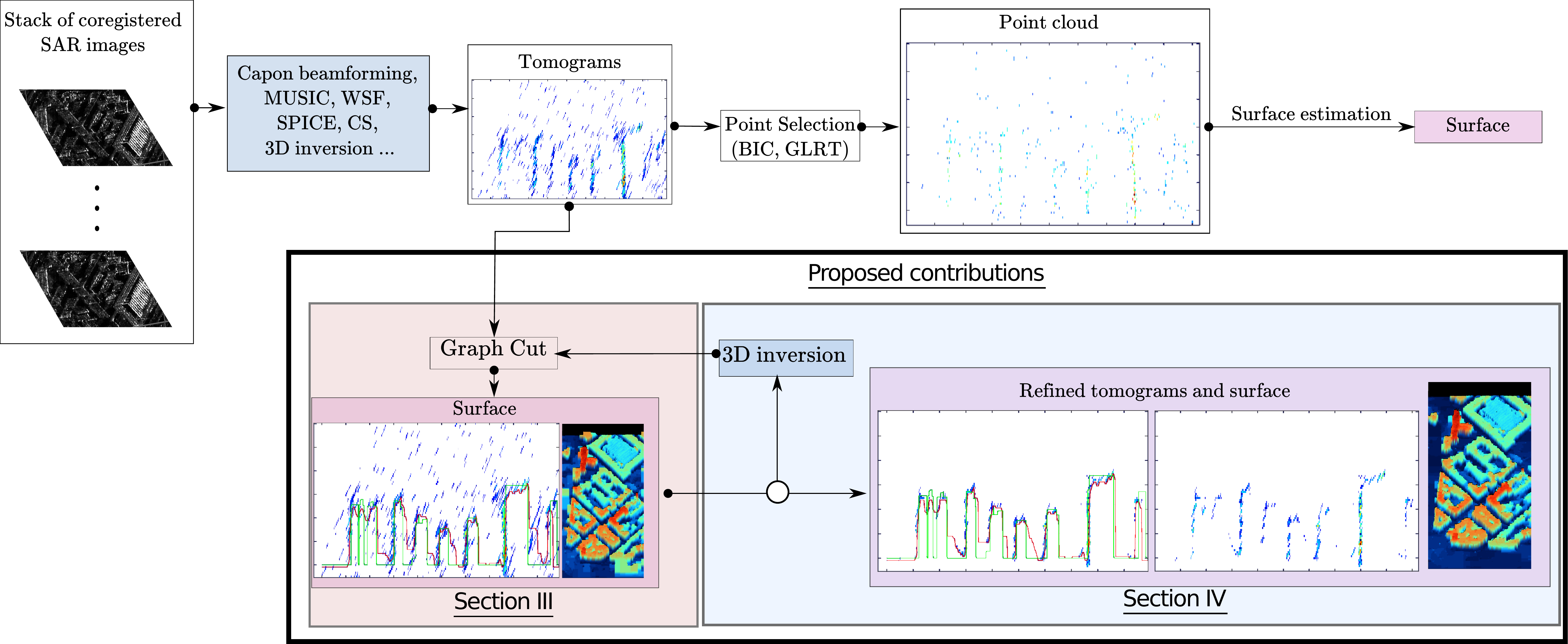}
	\caption{Overview of the article and proposed workflow. Starting from a stack of coregistered SAR images, different SAR tomographic techniques can be used to retrieve a 3-D estimation of the reflectivity. As the reconstructions are generally corrupted with noise and/or outliers, urban surface segmentation algorithms are generally applied on a point cloud extracted from the tomograms. The number of points is obtained either by minimizing an information criterion such as BIC or AIC or through hierarchical hypothesis testing. Rather than extracting a set of points, we estimate directly the urban surface on the tomograms by a graph-cut method (see section 3). We also derive an alternating scheme that iteratively estimates the urban surface and uses it to improve the tomographic inversion (see section 4).}
	\label{fig:workflow}
\end{figure*}

The proposed contributions presented here are then a new graph-cut based segmentation algorithm to retrieve urban surface from any kind of tomographic reconstruction and an improved alternate algorithm to obtain both a 3-D estimation of the reflectivity and the urban surface. The contributions and the workflow are summarized in Fig. \ref{fig:workflow}.
The structure of the paper is as follows. We first describe the multi-baseline SAR signal model and state-of-the-art methods in urban SAR tomography. Then in Section \ref{sec:gc} we introduce a graph-cut based urban
surface segmentation method.  In Section \ref{sec:alt}, we propose a method that combines 3-D reconstruction and surface segmentation. We illustrate the efficiency of the segmentation depending on the tomographic reconstruction algorithm used on a stack of 40 images of
the city of Paris obtained by the TerraSAR-X satellite in the last section. Results obtained with the REDRESS algorithm are also presented on the same dataset. 

\section{State of the art in SAR tomography}
\subsection{Signal Model}
\begin{figure}[h]
	\centering
	\includegraphics[width=\linewidth]{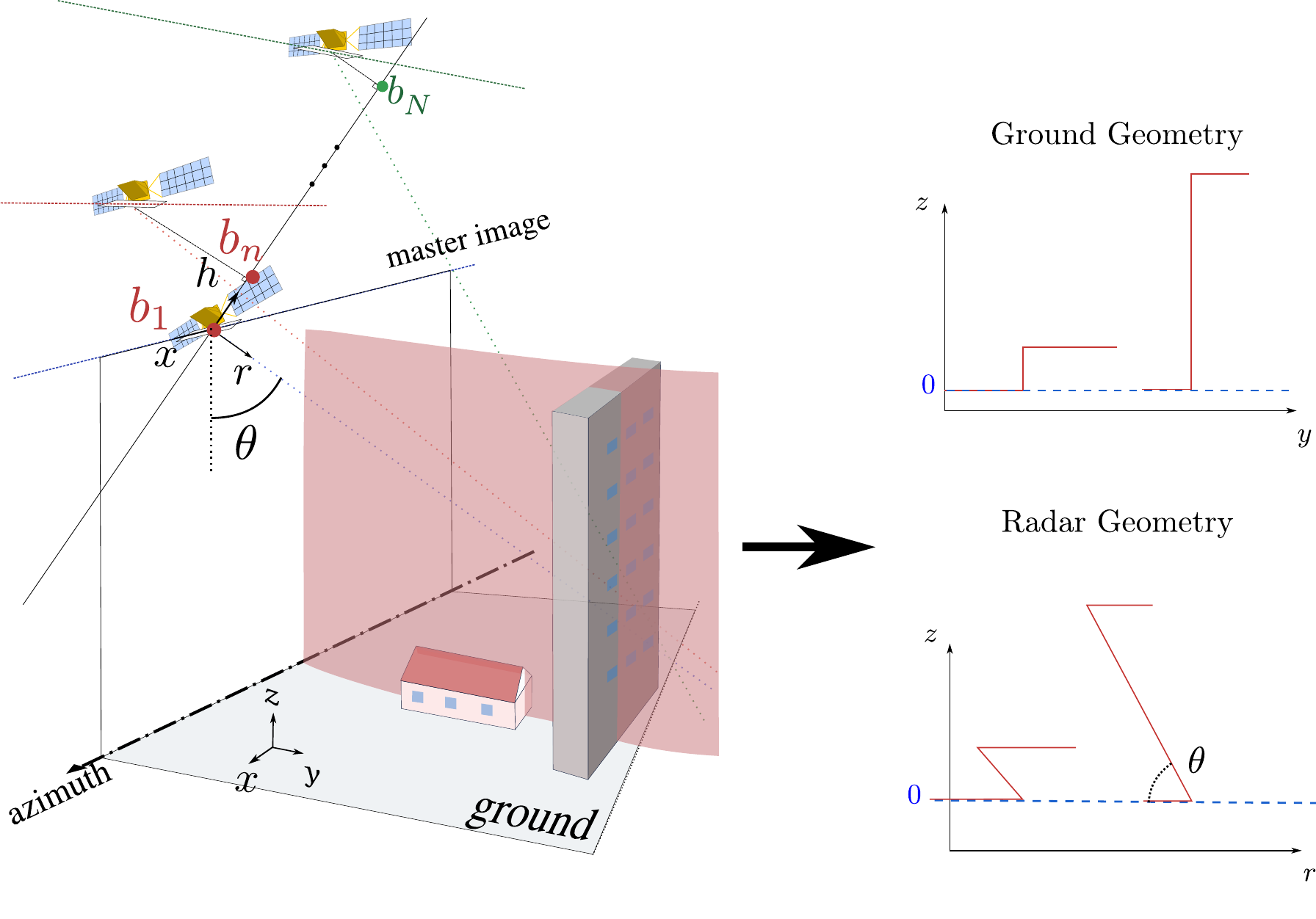}
	\caption{Geometry of the 3-D scene. Two geometric configurations can be used to describe
		the scene: the ground geometry associated with the coordinate system $(x,y,z)$
		and the radar geometry associated with the coordinate system $(x,r,z)$. On the
		left hand side, the different coordinates system are illustrated on a simple urban scene model. On the
		right hand side, a sketch of a tomographic profile corresponding to the red slice in each geometry.}
	\label{fig:tomo_geo}
\end{figure}

SAR tomographic signal models are generally written at a given SAR resolution cell.
However, to be able to express a spatial regularization, we prefer to define the reflectivity distribution in the 3-D space.
A SAR tomographic stack is obtained from $N$ Single Look Complex (SLC) SAR
images by spatial co-registration on a master image. The geometry of the scene is
depicted in Fig. \ref{fig:tomo_geo}. The complex amplitude at a given resolution cell of the images can be modeled as
the sum of the back-scattered amplitudes produced by all scatterers that are seen within that
same radar resolution cell. For the $n$-th image, the complex amplitude of the pixel at the
azimuth-range position $(x',r')$ is:
\begin{multline}
\underline{v}_n(x',r') = 
\iiint \underline{f}(x' - x,r' - \rho_{n;y,z}) \underline{u}(x,y,z)   
\cdot\text{exp}\left(-\frac{4\underline{j}\pi}{\lambda}
\rho_{n;y,z}+\underline{j}\varphi_{\text{atmo}}\right)\!\mathrm{d}x\,\mathrm{d}y\,\mathrm{d}z
+ \underline{\epsilon}(x',r')
\label{eq:tomo_principe}
\end{multline}%

where complex-valued variables are underlined for clarity, the variables $x,y$ and
$z$ denote the coordinates in the ground geometry whereas
$x'$ and $r'$ correspond to the coordinates along the azimuth and range axes. 
The distance $\rho_{n;y,z}$ is the distance of a scatterer at location $(x,y,z)$ to the closest point of the $n^{th}$ sensor trajectory, and
$\underline f$ represents the
Point Spread Function (PSF). $\varphi_{\text{atmo}}$ is the phase noise due to the atmospheric phase screen. The random variable $\epsilon(x',r')$ models the thermal noise (modeled as a white additive Gaussian process).

Each image is acquired from a slightly different angle at each pass of the
sensor. This angular diversity induces a different distance $\rho_{n;y,z}$ to each
antenna.
Upon proper sampling, co-registration and spectral apodization to reduce sidelobes,
the PSF can be roughly approximated by a 
2-D Dirac, leading to the following simplification:
\begin{align}
\underline{v}_n(x',r') = 
&\iint_{(y,z)\in\Delta_{r'}} \underline{u}(x',y,z) . \text{exp}\left(-\frac{4\underline{j}\pi}{\lambda}
\rho_{n;y,z}+\underline{j}\varphi_{\text{atmo}}\right)\,\mathrm{d}y\,\mathrm{d}z +
\underline{\epsilon}(x',r')\,,
\label{eq:tomo_principe2}
\end{align}
where the integration is carried out over points $(y,z)\in\Delta_{r'}$, i.e., points such that $\rho_{n;y,z}=r'$.
The tomographic stack can be pre-processed as a multi-baseline interferometric stack in order to directly relate the phases to heights.
Under some classical approximations (see
 \citep{Fornaro_tomo_principe} for details), and after removal of the atmospheric phase screen,
(\ref{eq:tomo_principe2}) can be written:
\begin{align}
\underline{v}_n(x',r') = 
\iint_{(y,z)\in\Delta_{r'}}
\!\!\!\!\!\underline{u}(x',y,z)\, \text{exp}\,\bigl(-\underline{j} \xi_n
z \bigr)\,\mathrm{d}y\,\mathrm{d}z + \underline{\epsilon}\,.
\label{eq:tomo_principe3}
\end{align} 
The $r'$-th radar resolution cell is defined by:\\
$\Delta_{r'}=\{\;(y,z)\;|\;r'-\delta_{\text{range}}/2\leq\rho_{y,z}\leq
r'+\delta_{\text{range}}/2\;\}$, with $\delta_{\text{range}}$
the step in range direction. The set $\Delta_r$ is thus the
extension of the radar resolution cell along its elevation
direction $h$ (\textit{cf. Fig. \ref{fig:tomo_geo}}).
The parameter $\xi_n = \frac{4 \pi
	b_{n}}{\lambda r' \sin \theta}$ is the spatial impulse associated to the
sampling of the scene for each baseline, $b_{n}$ is the
$n$-th baseline, $\theta$ the incidence angle of the master sensor and $\lambda$ the radar wavelength.


\subsection{Covariance based techniques}

In most of SAR tomographic approaches, the inversion is performed resolution-cell by
resolution-cell, keeping the original radar geometry. A 3-D representation of the scene is then
obtained by merging all 1-D inversions. Throughout this
section, we will consider the complex amplitudes measured at
given pixel.
We denote by $\Vu v\in\mathbb C^N$ the vector formed by the complex
amplitudes observed at the pixel of interest on each of the $N$
images. This signal results from the back-scattering produced by
scatterers with reflectivity $\Vu u\in\mathbb C^D$, each scatterer
located at the same range but at $D$ different elevations along the
$z$ axis. By discretization of equation (\ref{eq:tomo_principe3}), the
model of the measurement corresponds to the following linear model
under additive noise:
\begin{align}
\Vu v = \Vu A(\V z) \cdot\Vu u + \Vu\epsilon
\label{eq:tomo_1D}
\end{align}
where matrix $\Vu A(\V z) \in \mathbb{C}^{N\times D}$ is called
the sensing or steering matrix and consists of the
concatenation of the $D$ steering
vectors associated to each scatterer of the radar cell. It
depends on the sampling of elevations $\V h$.
A steering vector $\Vu
a_{1\leq k \leq D}\in \mathbb{C}^{N} $ is defined by
\begin{align}
\Vu a_k = \begin{bmatrix}
\exp(-\underline j \xi_1
z_k) \cdots \exp(-\underline j \xi_N
z_k)
\end{bmatrix}^{\text{t}}
\label{eq:steering_vector}
\end{align}

Many estimators exist in the field of spectral analysis to invert
(\ref{eq:tomo_1D}). They all require an estimate of the
covariance matrix $\Vu R = \mathbb E[\Vu a_k \Vu a_k^{\text{H}}]
\in \mathbb{C}^{N\times N}$ at the pixel of interest. As
discussed in the introduction, estimating this covariance
matrix by local averaging is tricky in heterogeneous regions
(high risk of mixing signals corresponding to very different
scatterers configurations) and when considering large stacks
(the sample covariance matrix is singular unless a least $N$
samples are averaged, which might represent more than one
hundred pixels).

The classical beamforming and the Capon beamforming are two
popular nonparametric estimators that lead to good results in
the SAR tomographic representation of continuous reflectivity
profiles such as in forests or ice. Due to
their simplicity and computational efficiency, they can
provide a quick overview of urban 
landscape. However, as classical beamforming does not provide a side-lobe
suppression, outliers are likely to occur on urban areas where
dynamic range is much larger than it is in forests or ice regions.	
Several methods consider a limited number of scatterers $D <
N$ (sparse spectral estimators): MUSIC \citep{MUSIC} and the
related WSF techniques \citep{Viberg,hue_laurent}. 
MUSIC is built on the consideration that, when there are $D<N$
scatterers located at $D$ given locations, the observed vector of
complex amplitudes is located close to the sub-space spanned by the
corresponding $D$ steering vectors. Hence, the $N-D$ lowest
eigenvalues of the covariance matrix span the complementary orthogonal
sub-space (the so-called noise subspace).
%
In WSF this idea is refined by including the distribution of the
noise eigenvectors of the empirical covariance matrix. When only the noise subspace is considered this approach is called NSF for Noise Subspace Fitting. Beside an estimation of
the covariance matrix, those techniques also require an estimation of the number
of scatterers present in each radar cell.

The recent SPICE method \citep{spice} is a fully non-parametric sparse algorithm
based on the minimization of a covariance fitting criterion between the estimated
covariance matrix and its theoretical expression according to
(\ref{eq:tomo_1D}). This algorithm achieves a very good performance when the
covariance matrix is correctly estimated and when the discrete model is
respected.

\subsection{Regularized inversion}
More suited to dense urban areas than methods based on covariance matrix
analysis, CS has proved to be able to achieve super-resolution
in the scatterers unmixing \citep{Zhu_CS_intro,Schirinzi_CS_intro}. Since no covariance matrix
estimation is required, this method avoids the resolution-loss
implied by filtering-based covariance estimation. CS performs
the inversion of (\ref{eq:tomo_1D}) under a sparse prior:
\begin{align}
\Vhu u = \underset{\Vu u}{\text{argmin}}\;\; || \Vu P \Vu u - \Vu v ||_2^2
+ \mu ||\Vu u||_1
\label{eq:CS}
\end{align}
The parameter $\mu$ balances the importance of the sparsity prior with
respect to the data fidelity. $\Vu P$ is a block diagonal matrix
with each block being the steering matrix $\Vu A$ associated to the pixels in the SAR images.

In urban areas where bright scatterers
can be located near empty radar cells this parameter tuning
can be challenging (i.e., require \emph{local} tuning).
Although being parametric, parameter $\mu$ is not directly
expressed in terms of the number of scatterers in the cell.

Instead of performing the inversion in the radar geometry,
the 3-D inversion \citep{3Dinv} is performed directly in ground geometry to allow the use 
of geometrical priors.
However if only a sparse prior is used we obtain the equivalent problem:
\begin{align}
\Vhu u = \underset{\Vu u}{\text{argmin}} || \Vu \Phi\, \Vu u - \Vu v ||_2^2 +
\mu||\Vu u||_1
\label{eq:3Dinv}
\end{align}
where the linear operator $\Vu \Phi$ maps the complex-valued
reflectivities of the scatterers sampled in ground geometry to the
measurements (complex amplitudes on the SAR antennas):
\begin{align}
\label{eq:phi_def}
\Vu \Phi_{k,\ell} &= \begin{cases}
0&\text{ if }\rho_{1;y_\ell,z_\ell}\notin
[r_k'-\frac{\delta_{\text{range}}}{2},r_k'+\frac{\delta_{\text{range}}}{2}]\\
\exp(-\underline j \varphi_\ell)&\text{otherwise,}
\end{cases}
\end{align}
with $\varphi_\ell$ the phase model given either by equation (\ref{eq:tomo_principe}):
$\varphi_\ell = 4 \pi \rho_{n;y_\ell,z_\ell}/\lambda$, or by
equation (\ref{eq:tomo_principe2}): $\varphi_\ell = \xi_n z_\ell$.

Equations (\ref{eq:CS}) and (\ref{eq:3Dinv}) differ only from their inverse operator
and thus in the final geometry in which $\Vu u$ is described.
We chose the second framework to perform the sparse inversion as it can be more
easily extended to locally adapt the weight of the sparsity
prior according to the urban surface retrieved by the
graph-cut segmentation technique described next.

	\section{Graph-cut based surface segmentation}
	\label{sec:gc}
	Starting from the tomographic reconstruction obtained with one of the
	methods described in the previous section (a 3-D volume $u(x,y,z)$),
	we aim to recover the urban surfaces (ground, building facades,
	roofs). Following a typical approach in computer vision for surface
	reconstruction, we formulate the problem as an energy minimization
	problem. We seek a surface $\mathscr{S}$ corresponding to an elevation
	map: $(x,y)\mapsto z=\mathscr{E}(x,y)$ that both
	fits well the reconstructed tomographic volume and that is smooth.
	We first formulate a cost function that captures these two properties,
	then we describe an efficient graph-based algorithm to perform the
	minimization of the cost function.
	
	\subsection{Definition of the cost function}
	\begin{figure}[!h]
		\centering
		\includegraphics[width=.8\linewidth]{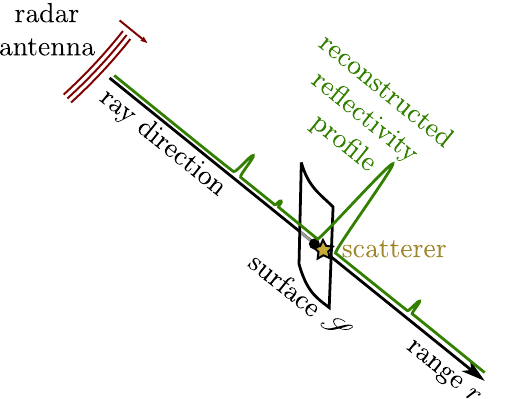}
		\caption{We seek a surface $\mathscr{S}$ that, for each ray, is
			close the scatterer(s) found along the ray.}
		\label{fig:raysurfaceinter}
	\end{figure}
	The first component of the cost function favors surfaces that are
	faithful to the reconstructed tomographic volume. We seek surfaces
	such that, when considering a given ray direction in 3-D space, the
	scatterer encountered along the ray falls close to the ray-surface
	intersection, see Fig.\ref{fig:raysurfaceinter}. The reflectivity
	profile along the ray may display several local maxima due to
	residual sidelobes after the tomographic inversion. Rather than
	detecting these maxima and deciding for the most meaningful maximum,
	we consider that a satisfying location of the surface is a location
	such that the reflectivity profile is split into two well-balanced
	halves. We define
	the cumulative reflectivity $C^-(r_s)$ from the antenna to the surface
	$\mathscr{S}$ and the cumulative reflectivity $C^+(r_s)$ from the
	surface to the maximum range: 
	\begin{align}
	C^-(r_s)&=\int_{r_{\text{min}}}^{r_s} |u(r)|\,\text{d}r,\\
	C^+(r_s)&=\int_{r_s}^{r_{\text{max}}} |u(r)|\,\text{d}r\,,
	\end{align}
	where $r_s$ is the range of the surface, i.e., the distance from the
	radar to the surface, in the direction of the ray. If the surface is
	such that $C^-(r_s)<C^+(r_s)$, then it is too close to the radar: most
	of the reflectivity of the scatterers encountered along the ray is
	located beyond the surface. Conversely, if $C^-(r_s)>C^+(r_s)$, the surface is
	too far from the radar: scatterers accounting for most of the
	reflectivity are located before the surface. The imbalance
	$C^-(r_s)-C^+(r_s)$ is therefore an indication of bad surface
	localization. In order to favor surfaces that are located close to the
	position of equilibrium, we define the penalty:
	\begin{align}
	D(r) = \int_{r_{\text{min}}}^r \left[C^-(r_s)-C^+(r_s)\right]_+\text{d}r_s+
	\int_{r}^{r_{\text{max}}} \left[C^+(r_s)-C^-(r_s)\right]_+\text{d}r_s\,,
	\label{eq:dist}
	\end{align}
	where the notation $[\cdot]_+$ denotes the positive part:
	$\forall w,\,[w]_+=\max(w,0)$. The term
	$\left[C^-(r_s)-C^+(r_s)\right]_+$ in the first integral of equation
	(\ref{eq:dist}) is non-zero only if the distance $r_s$ is larger than
	the distance of equilibrium $r_{\text{equi}}$ (where $r_{\text{equi}}$
	is such that $C^+(r_{\text{equi}})=C^-(r_{\text{equi}})$). Then, if
	$r>r_{\text{equi}}$, the first integral equals
	$\int_{r_{\text{equi}}}^r
	\left(C^-(r_s)-C^+(r_s)\right)\text{d}r_s$. Conversely, the second
	integral in (\ref{eq:dist}) is non-zero only if the distance $r_s$ is
	smaller than the distance of equilibrium $r_{\text{equi}}$. It is then
	equal to $\int_r^{r_{\text{equi}}}
	\left(C^+(r_s)-C^-(r_s)\right)\text{d}r_s$. $D(r)$ is thus a function
	that \emph{monotonically increases with the distance} $|r-r_{\text{equi}}|$
	and that is minimal and equal to zero when $r=r_{\text{equi}}$.
	
	\medskip
	The second component of the cost function guarantees that the
	segmented surface is smooth. To prevent the surface from oscillating
	in order to pass through the position of equilibrium $r_{\text{equi}}$
	for each ray, we penalize the area $\mathscr{A}(\mathscr{S})$ of the
	surface. In order to favor surfaces with horizontal or vertical
	parts, we suggest measuring the area with respect to the $\ell_1$ distance
	(i.e., Manhattan distance $\|\V p\|_1=|p_x|+|p_y|+|p_z|$).
	
	\medskip
	To summarize, we suggest defining the segmentation as the surface
	$\mathscr{S}$ that is a solution to the following variational problem:
	\begin{align}
	\min_{\mathscr{S}} \;\;\int_{\text{ray}\in\mathscr{R}}
	D_{\text{ray}}(r_{\text{ray}\to\mathscr{S}})\,\text{d}\mathscr{R}
	\;\;+\;\; \beta \,\mathscr{A}(\mathscr{S})\,,
	\label{eq:variational}
	\end{align}
	where $\mathscr{S}$ is required to be representable as an elevation
	map $\mathscr{E}(x,y)$ (formally, there exist a function $\mathscr{E}:\;(x,y)\mapsto
	\mathscr{E}(x,y)$ such that $\mathscr{S}$ be the
	boundary of the epigraph of $\mathscr{E}$). To prevent from
	introducing too many notations, we denote 'ray' for the generic
	definition of a ray in an adequate parameterization (a line in 3-D
	space), $\mathscr{R}$ represents the set of all rays,
	$r_{\text{ray}\to\mathscr{S}}$ is the distance from the radar to the
	surface $\mathscr{S}$ along the direction defined by 'ray',
	$D_{\text{ray}}$ is the penalty defined by equation (\ref{eq:dist})
	for the direction specified by 'ray'. Finally, $\beta$ is a parameter
	that balances the fidelity to the tomographic reconstruction and the
	spatial smoothness of the surface.


	\subsection{Graph-cut algorithm for minimization}
	
	\begin{figure*}[t]
		\centering
		\includegraphics[width=.9\textwidth]{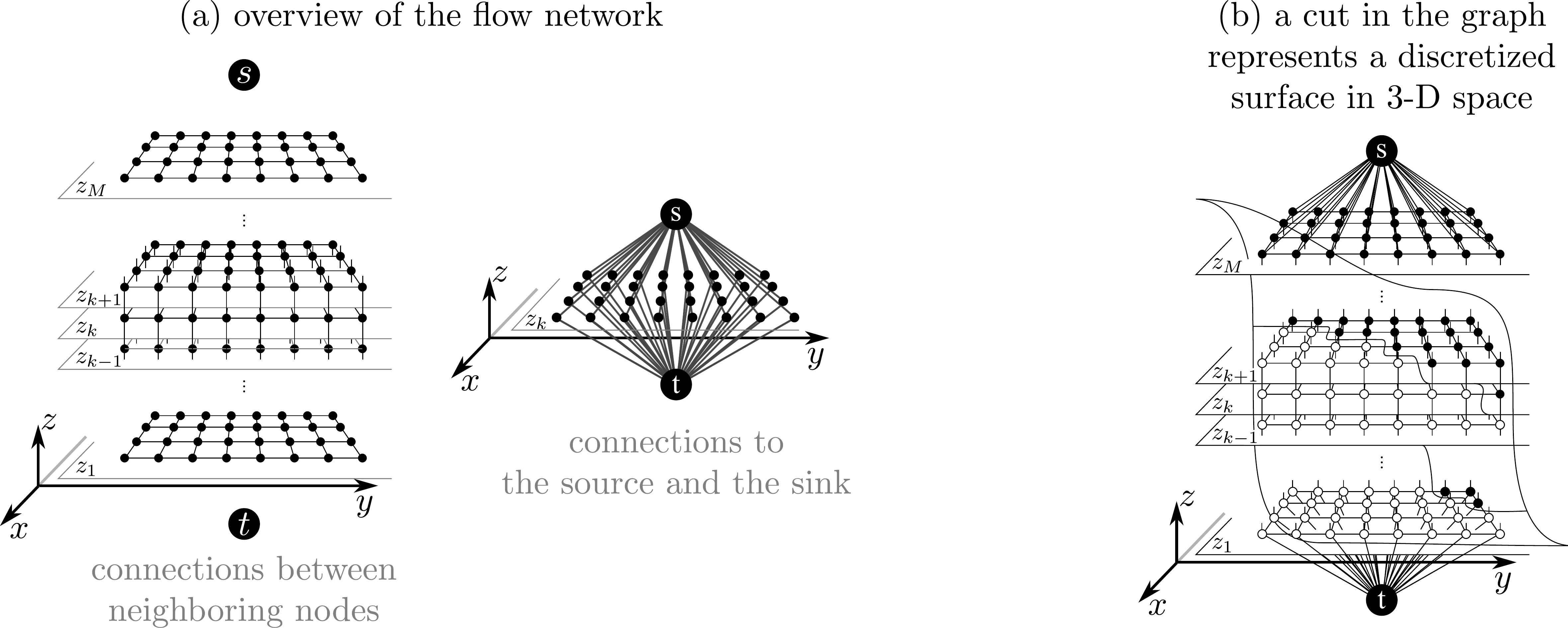}
		\caption{Representation of the topology of the flow network: (a) a node
			represents a 3-D $(x,y,z)$ location in ground geometry, each node
			is connected to its 6 closest neighbors and also to the source s
			and to the sink t; (b) a cut separates the graph into
			two disconnected sub-graphs, it represents a discretized version
			of the segmented surface $\mathscr{S}$.}
		\label{fig:graph}
	\end{figure*}
	
	The variational problem (\ref{eq:variational}) is very challenging to
	solve. We show in this paragraph that, after discretization of the
	surface and of the set of rays, it can be transformed into a minimum cut
	problem on a particular graph. By computing the minimum cut using
	available efficient graph-cut libraries, we obtain a fast method to
	solve the surface segmentation problem.
	
	The surface is represented by an elevation map
	$\mathscr{E}:\;(x,y)\mapsto z=\mathscr{E}(x,y)$
	(which guarantees that it is representable as an elevation map). The
	horizontal location $(x,y)$ and the elevation $z$ are discretized. To
	make an easier connection between the elevation map and the surface it
	defines, we consider the layer cake decomposition of the
	elevation. With this decomposition, a discrete elevation map
	corresponds to a binary volume (a discrete version of the epigraph of
	$\mathscr{E}$) and the boundary in that volume defines the discrete
	surface.
	
	We build a graph as depicted in Fig. \ref{fig:graph}, with a node to
	represent each voxel of the binary volume of $\mathscr{E}$. Two
	special nodes, called the source (denoted 's') and the sink (denoted
	't') are added in order to simulate a flow from the source to the
	sink. Nodes are connected together by directed edges with specific
	capacities and a flow is said to be admissible if and only if the flow
	along each edge is non negative and smaller or equal to the edge
	capacity, and there is no flow accumulation/creation at nodes (except
	at the source and at the sink). By
	the max-flow min-cut theorem, algorithms that identify the maximum
	admissible flow on the graph can also identify the minimum cost cut among
	all possible cuts in the graph\footnote{the cost of a cut is the sum
		of the capacities of all edges cut that are directed from a node in
		the source partition to a node in the sink partition}, see for example
	\citep{boykov2004experimental}. During the graph construction, by creating
	edges with well-chosen capacities, we can make the cost of any cut
	exactly match the cost of the corresponding surface in the variational
	formulation (\ref{eq:variational}).
	
\begin{figure*}[t]
		\centering
		\includegraphics[width=\linewidth]{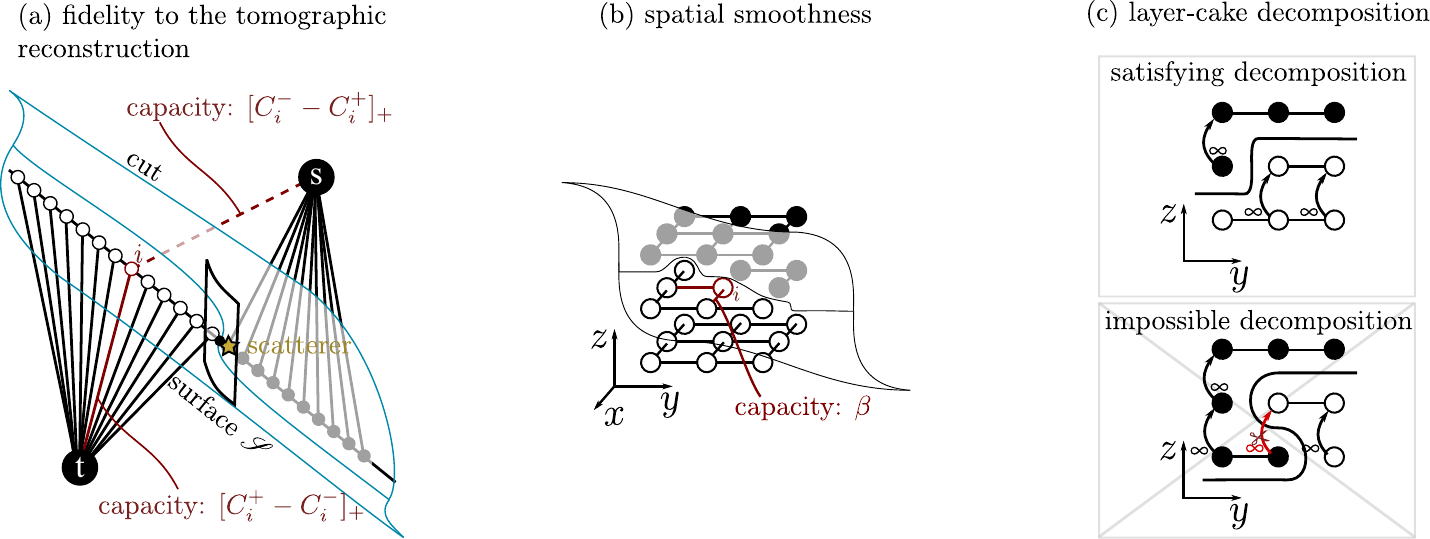}
		\caption{The capacities of the edges are chosen so that the cost of
			the cut corresponds to the energy of the surface. (a) the fidelity
			to the tomographic reconstruction is enforced via edges
			originating from the source or leading to the sink. (b) the
			spatial smoothness of the surface is obtained by adding
			bi-directional edges between neighboring nodes in the $x$ and $y$
			directions. (c) to prevent the cut from severing twice a column of
			nodes along the $z$ direction, ascending edges with infinity
			capacity are added. These edges are counted in the total cost of
			the cut only when they go down-stream: from the partition
			containing the source to the partition containing the sink.}
		\label{fig:raysurfaceinter2}
	\end{figure*}

	To represent the first term in
	equation (\ref{eq:variational}), we substitute $D_{\text{ray}}$ with
	its definition in equation (\ref{eq:dist}):
	\begin{multline}
	\int_{\text{ray}\in\mathscr{R}}
	D_{\text{ray}}(r_{\text{ray}\to\mathscr{S}})\,\text{d}\mathscr{R}=
	\int_{\text{ray}\in\mathscr{R}}
	\int_{r_{\text{min}}}^{r_{\text{ray}\to\mathscr{S}}} \left[C_{\text{ray}}^-(r_s)-C_{\text{ray}}^+(r_s)\right]_+\text{d}r_s\,\text{d}\mathscr{R}\\+\int_{\text{ray}\in\mathscr{R}}
	\int_{r_{\text{ray}\to\mathscr{S}}}^{r_{\text{max}}}
	\left[C_{\text{ray}}^+(r_s)-C_{\text{ray}}^-(r_s)\right]_+\text{d}r_s\,\text{d}\mathscr{R}\,.
	\label{eq:Dray2}
	\end{multline}
	Each of the two terms correspond to summations over a half-space whose
	boundary is $\mathscr{S}$: the half-space that contains the radar and
	the half-space with the farther ranges, respectively. We add an
	edge directed from the source to node $i$, the node that represents the 3-D
	position $(x_i,y_i,z_i)$ and that is located at the distance $r_i$
	from the radar antenna. The capacity\footnote{note that an edge with
		zero capacity can be suppressed because it carries no flow and has
		no contribution to the cost of the cuts} of this edge is set to
	$\left[C_i^-(r_i)-C_i^+(r_i)\right]_+$, where $C_i^-$ and $C_i^+$ are
	the cumulative reflectivities computed along the ray directed from the
	radar through the point of coordinates $(x_i,y_i,z_i)$. Another
	directed edge is added from node $i$ to the sink, with capacity
	$\left[C_i^+(r_i)-C_i^-(r_i)\right]_+$. To separate the graph into two
	parts by a cut, some edges must be severed (unless the cut passes
	through the distance of equilibrium $r_{\text{equi}}$) and the sum of
	the capacities of those edges corresponds to a discretization of
	equation (\ref{eq:Dray2}), see Fig. \ref{fig:raysurfaceinter2}(a).

	Additional edges are created to account for the regularization term
	$\beta\,\mathscr{A}(\mathscr{S})$: bi-directional edges between pairs of
	nodes that are direct neighbors in the $x$ or $y$ directions, with
	capacity $\beta$, see
	Fig. \ref{fig:raysurfaceinter2}(b). Finally, ascending edges with
	infinite capacity are included between neighboring nodes in the $z$
	direction. These edges are necessary to guarantee that the cut defines
	a surface that is representable by an elevation map, see
	\ref{fig:raysurfaceinter2}(c). Similar edges are added in Ishikawa's
	graph construction that is also based on the layer-cake decomposition \citep{Ishikawa}.
	
	In our implementation, we computed efficiently the summations along
	the rays by resampling the reconstructed tomographic volume in ray
	geometry so that sums could be carried out along columns in this new
	geometry. For the construction of the graph and the computation of the
	minimum cut, we used the graph-cuts
	library by Boykov and Kolmogorov \citep{boykov2004experimental}.
	\revc{The number of nodes in the graph is equal to the number of 
	voxels in the estimated volume, \textit{i.e.}, the number
	of pixels in one SAR image times the number of heights considered along the vertical direction.
	The number of edges is proportional to the number of nodes.
	The maximum complexity of a cut in the graph is $O(EV^2C)$
	with $E$ being the number of edges, $V$ the number of vertices and
	$C$ the value of the cut, but the experimental complexity is almost linear in the number of vertices \citep{boykov2004experimental,Lobry_2016}. 
	On a computer with an Intel(R) Xeon(R) CPU E5-2698 v4 @ 2.20GHz with 20 cores and 120 GB of RAM, finding the minimum cut
	takes $43\,\text{s}$ when dealing with a volume of $8.96 \times 10^6$ voxels, for a typical value of the regularization parameter.
}

\section{Joint reconstruction and surface segmentation}
\label{sec:alt}

As mentioned in the introduction, the knowledge provided by the segmented urban
surfaces can help to improve the inversion. 
The reconstruction algorithm that can most readily be extended to
include segmented surfaces is the 3-D inversion method described in
equation (\ref{eq:3Dinv}).
Under the
assumption that the signal retrieved over urban areas is mainly constituted of
punctual bright points, sparsity may be an efficient enough prior to obtain
clean tomograms. Nonetheless, this implies that the tuning of the sparsity
parameter shall be done locally according to the position of the scatterers. In CS
for SAR tomography, the sparsity constraint is generally set locally in the
range and azimuth direction but constant for each radar cell. Here we propose to
use the 3-D information provided by the estimated surface to
go one step further and perform a spatially varying
penalization of the sparsity.

When applying CS or the 3D inversion, the sparsity parameter
$\mu$ is set 
proportional to the level of spurious elements in the reconstruction.
Generally $\mu$ is set according to the noise level \citep{Zhu_Robustness_CS}, but
as decorrelation mechanisms and side-lobes should also be discarded, the
knowledge of the sensor thermal noise may not be enough. Many SAR tomographic
algorithms propose to estimate the number of backscattering
elements in order to extract the
largest scatterers in each radar cell. This step cleans the
estimated tomograms from residual outliers, but is also a challenging task for
large multitemporal stacks in dense environments. Moreover the CS approach may
then loose one of its asset with respect to MUSIC or WSF if it also needs an
estimation of the number of targets.

Under the assumption that the location of the urban surface is known, the
sparsity parameter $\mu$ can be spatially tuned 
to lead to refined
tomograms. Even when the surface is roughly known, it provides information on
where the reconstructed signal should be located. 
In the proposed iterative algorithm, $\mu$ is computed as a function of the
distance to the surface in the 3-D space and the number of iterations:
\begin{align}
\mu_k(\V p,\mathscr{S}) = \mu_0 + \frac{b}{(n-1)^2}\biggl(\frac{k}{n -
	k}\;{d}(\V p,\mathscr{S})\biggr)^2
\label{eq:mu}
\end{align}
where $d(\V p,\mathscr{S})$ is the Euclidean distance from the point $\V p = (x,y,z)^T$ to the
estimated surface, $k$ is the current iteration and $n$ the total number of
iterations. We define by $\V \mu(\mathscr{S})\in \mathbb{R}^{N_x . N_y . N_z}$ the 3-D sparsity parameter map.
As the surface location estimation may be subject to errors in the first
iterations, it is important to avoid over-penalizing points moderately close to the
surface during the first reconstructions. This is why we multiply
the distance $d$ by a factor smaller than 1. As the number
of iterations increases, the reconstruction and thus the surface estimation should
be more accurate (and better in match) which suggests an
increase of the penalization of the distance from
a reconstructed voxel to the surface. $\mu_0 + b$ is then the desired minimal sparsity 
that need to be apply to voxels not on the surface.

The proposed iterative reconstruction and surface segmentation is summarized in
the REDRESS algorithm.
\vspace{.1cm}
\renewcommand{\thealgorithm}{}
\begin{algorithm}
	\vspace{.3cm}
	\caption{{ Alte{\bf R}nat{\bf E}d 3-{\bf D} {\bf
				RE}construction and {\bf S}urface
			{\bf S}egmentation (REDRESS)}}
	\begin{algorithmic}[1]
		\renewcommand{\algorithmicrequire}{\textbf{Input:}}
		\renewcommand{\algorithmicensure}{\textbf{Output:}}
		\REQUIRE
		\begin{itemize}
			\item[] $\Vu v$ \hfill (stack of SLC SAR images)
		\end{itemize}
		\ENSURE 
		\begin{itemize}
			\item[] $\hat{\Vu u}$ \hfill (3-D cube of complex reflectivities)
			\item[] \qquad$\mathscr{S}$ \hfill (urban surface)
		\end{itemize}
		\textit{Initialization} :\\
		\STATE{$k \leftarrow  0$}
		\WHILE {$k < n$}
		\STATE{$\hat{\Vu u}\gets
			\text{\texttt{3-D Inversion}}(\Vu v,\V \mu(\mathscr{S}))$}
		\STATE{$\mathscr{S} \leftarrow  \text{\texttt{graph cut}}(\Vhu u)$}
		\STATE{$k \leftarrow  k+1$}
		\ENDWHILE
		\RETURN $\hat{\Vu u}, \Vh{\mathcal S}$
	\end{algorithmic}
\end{algorithm}

The refined
tuning of the sparsity
according to the surface allows to considerably improve the
scatterers
localization and main lobe reduction. In some cases, however, the segmented surface
follows the lobe main
extension direction and is not as localized as would be
expected for a collection of point-like scatterers. In the
global reconstruction of the scene, most of the 
artifacts due to the TV penalization are suppressed after 10
iterations. The obtained surface is then very close to the
ground truth and provides the lowest error according to table
\ref{table:errors}.

\revc{As the 3-D inversion step estimates all the voxels at once, it
	can be quite computationally intensive to perform it multiple times.
	Using the same configuration as described previously, one iteration of this algorithm
	takes 2 min 49 s when estimating a 3-D scene of $8.96 \times 10^6$ voxels starting from
	a stack of $1.28 \times 10^5$ SAR pixels.
}

\section{Experiments}
\begin{figure}[!ht]
	\begin{minipage}[t]{.475\linewidth}
		\centering
		\subfloat[]{\includegraphics[width=.7\linewidth, height =
			5cm]{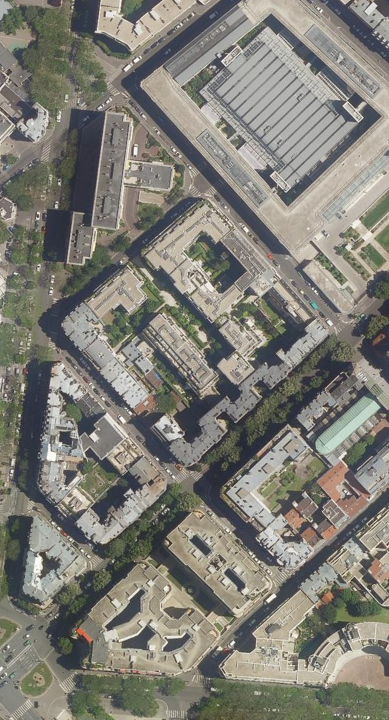}}\hfil
		\subfloat[]{\includegraphics[width=.7\linewidth, height =
			5cm]{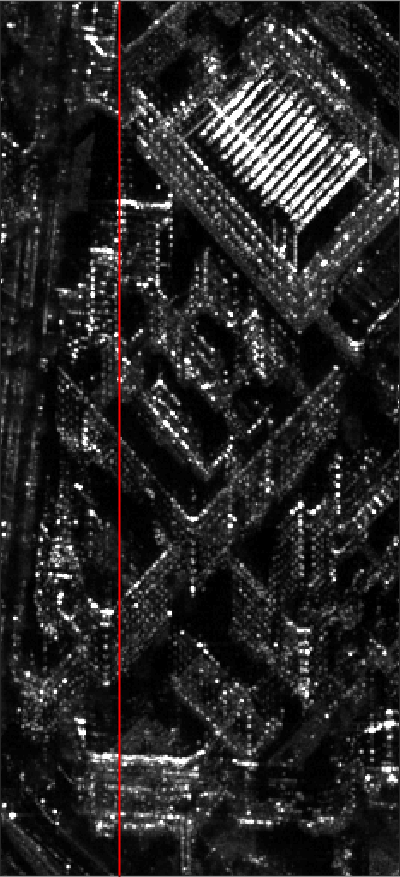}}\hfil
		\subfloat[]{\includegraphics[width=.7\linewidth, height =
			5cm]{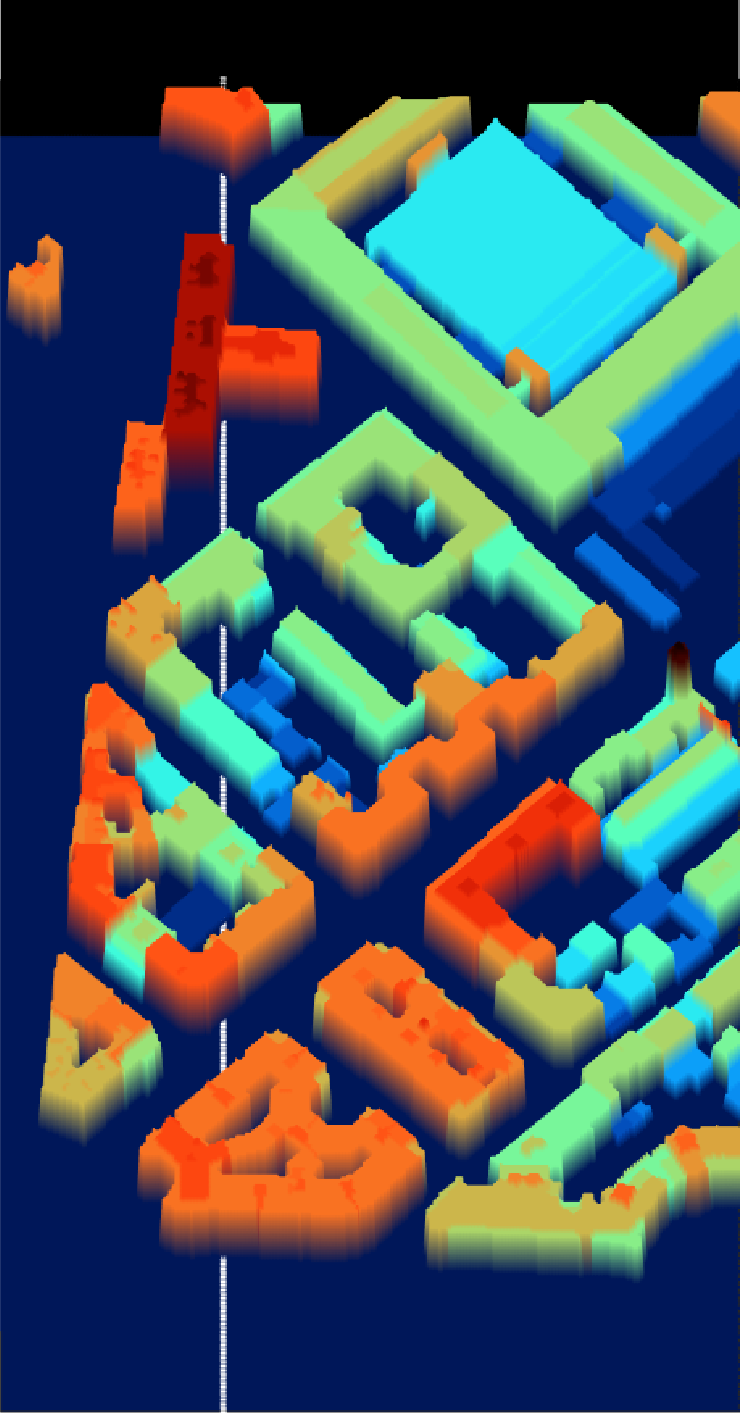}}\hfil
		\caption{Observed urban areas : optical image (a),
			temporal mean of the corresponding SAR image (b),
			and the 3-D model from IGN and Google Earth used as a ground truth.
			The red line in (b) and white one in (c) correspond to the slice shown in Fig.
			\ref{fig:slice}}
		\label{fig:data}
	\end{minipage}
	\begin{minipage}[t]{.05\linewidth}
		\hspace{\linewidth}
	\end{minipage}
	\begin{minipage}[t]{.475\linewidth}
	\centering
	\subfloat[]{\includegraphics[width=.7\linewidth, height =
		5cm]{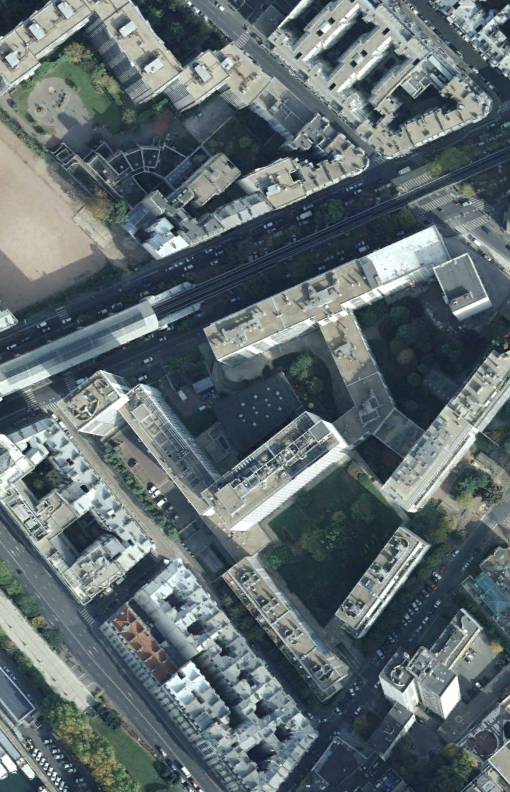}}\hfil
	\subfloat[]{\includegraphics[width=.7\linewidth, height =
		5cm]{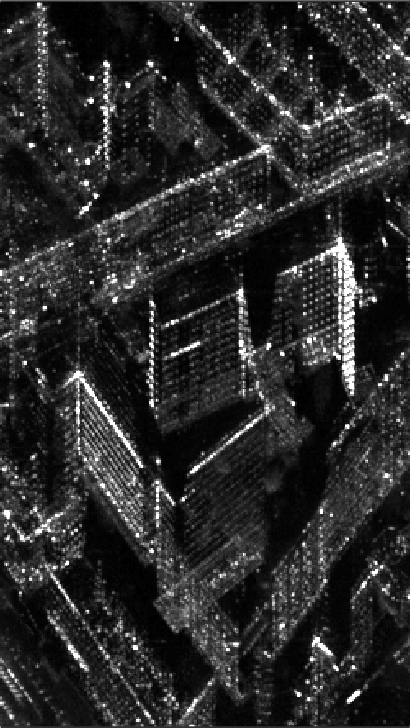}}\hfil
	\subfloat[]{\includegraphics[width=.7\linewidth, height =
		5cm]{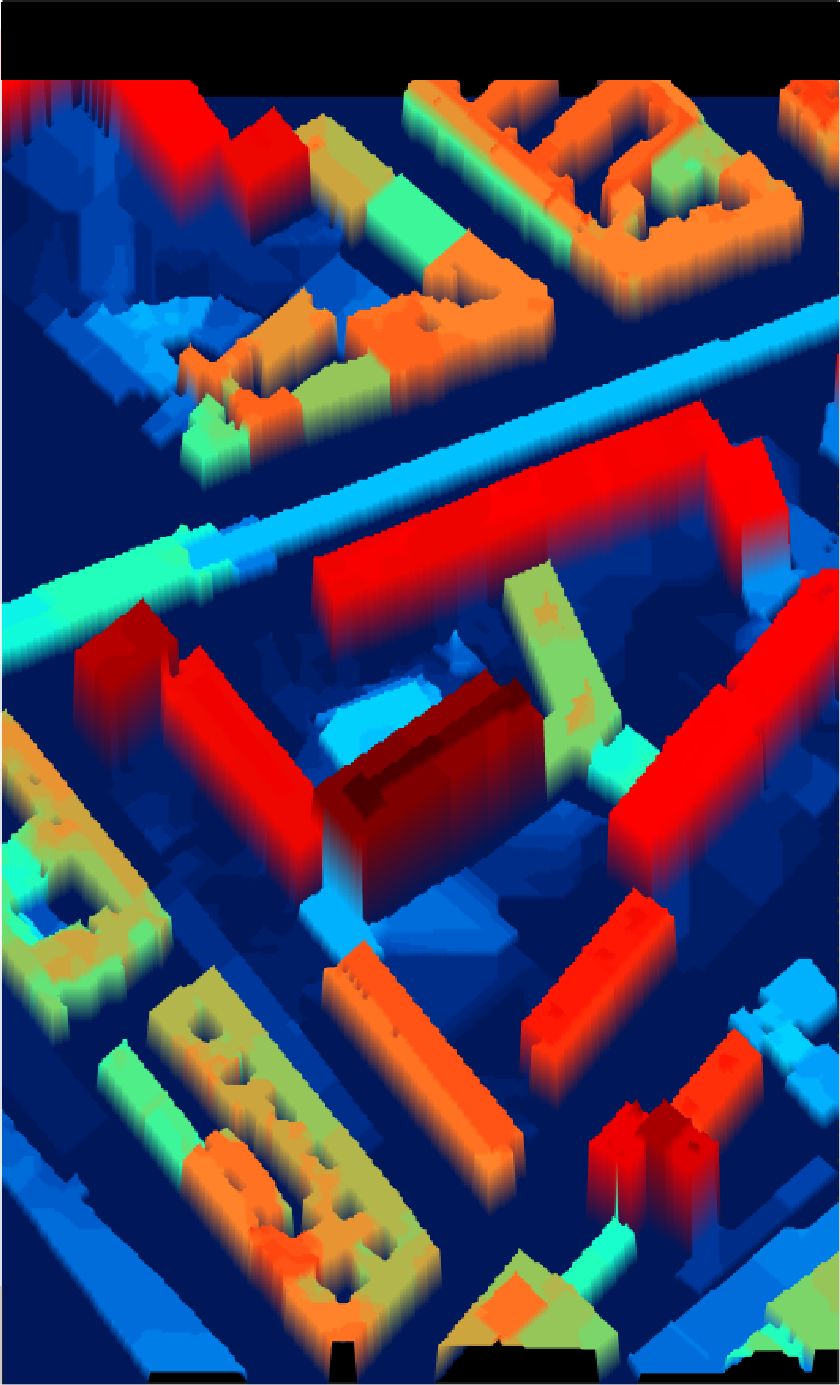}}\hfil
	\caption{Observed urban areas : optical image (a),
		temporal mean of the corresponding SAR image (b),
		and the 3-D model from IGN and Google Earth used as a ground truth.}
	\label{fig:data_b}
	\end{minipage}
\end{figure}

To validate both the generality of the segmentation method and its efficiency on real data we
present different experiments performed on \revb{two sets} of 40
TerraSAR-X images of a part of Paris, France. \revb{The first selected area corresponds to the French Ministry of Foreign
Affairs and to buildings in its neighborhood in the south west of the city. The building heights vary in a range from 10~m to 30~m.
 The
optical view of the scenes are presented Fig. \ref{fig:data} and Fig. \ref{fig:data_b}
side by side with the
temporal average of the SAR intensities. The second scene consists in buildings next to the \textit{rue de Grenelle}, with more diverse heights: the smallest building is only 8~m tall while the highest is close to 60~m.} \reva{
To evaluate our results, we built a ground truth from the building footprints and heights provided by the French geographical institute (IGN), with additional details (missing courtyard, rooftops irregularities) manually included from Google Earth\textsuperscript{TM} models.
} Different SAR tomographic
reconstruction methods introduced in section 2 (Capon Beamforming, MUSIC, WSF,
SPICE, CS and the 3-D inversion) are first applied on the slice represented
by the red line in Fig. \ref{fig:data} (b), then on the entire data sets. The
segmentation by graph-cut is then performed on all the scenes. \reva{The parameter $\beta$ is set to its optimal value with respect to a subset of the dataset presented in Fig. \ref{fig:data}.} The obtained surfaces are compared with the ground truth for each
tomographic estimator.
%

\begin{figure}[!ht]
	\centering
	\begin{minipage}{\linewidth}
		\subfloat[]{
			\begin{tikzpicture}
			\node (img)  {\includegraphics[width=.44\linewidth]{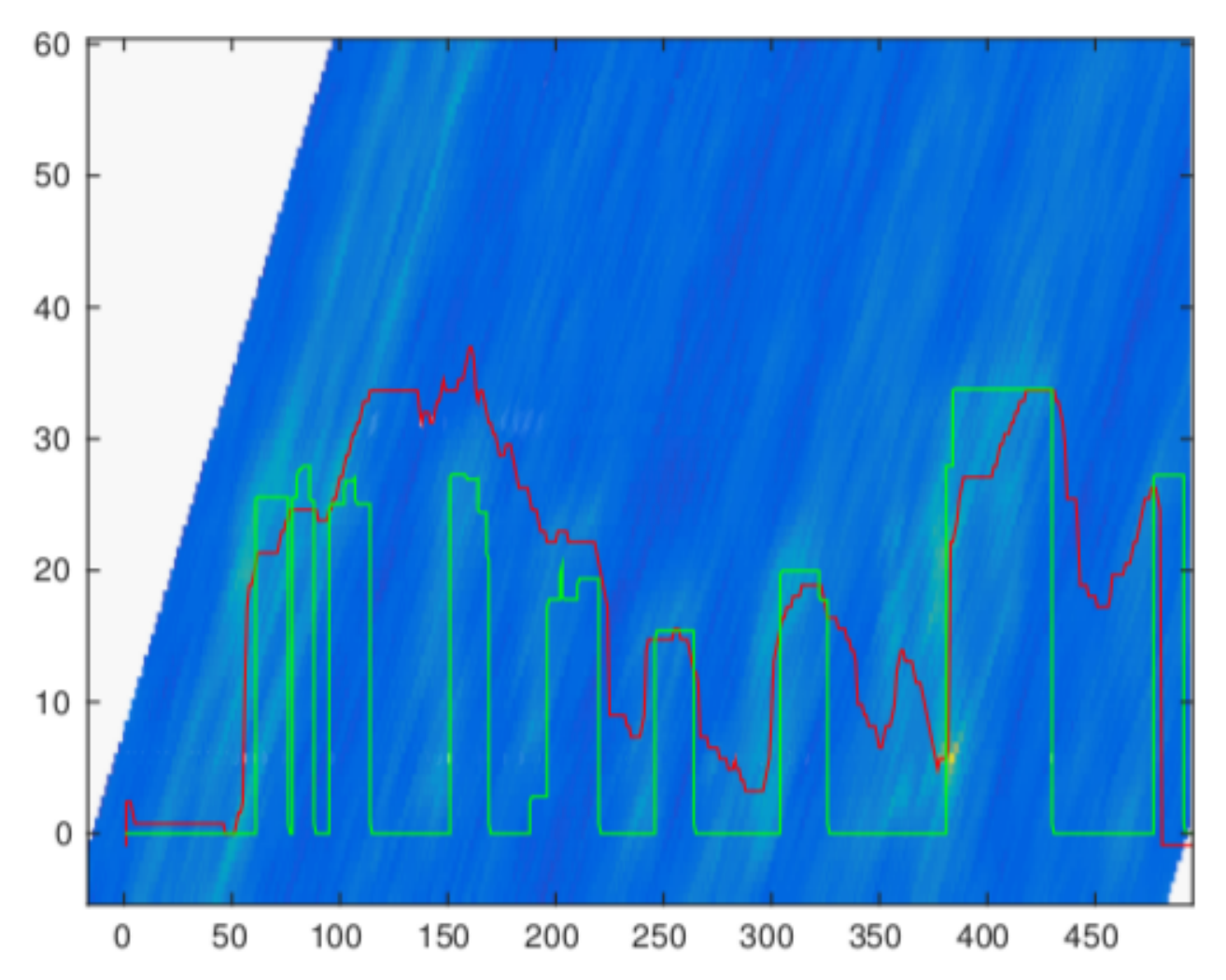}};
			
			\node[below=of img, node distance=0cm,xshift=0.cm, yshift=1.1cm, scale = .6
			,font=\color{black}] {y};
			
			\node[left=of img, node distance=0cm, rotate=0, scale = .6,
			anchor=center,yshift=0.45cm,xshift=1.7cm,font=\color{black}] {z};
			\end{tikzpicture}
		}\hfil
		\subfloat[]{
			\begin{tikzpicture}
			\node (img)  {\includegraphics[width=.44\linewidth]{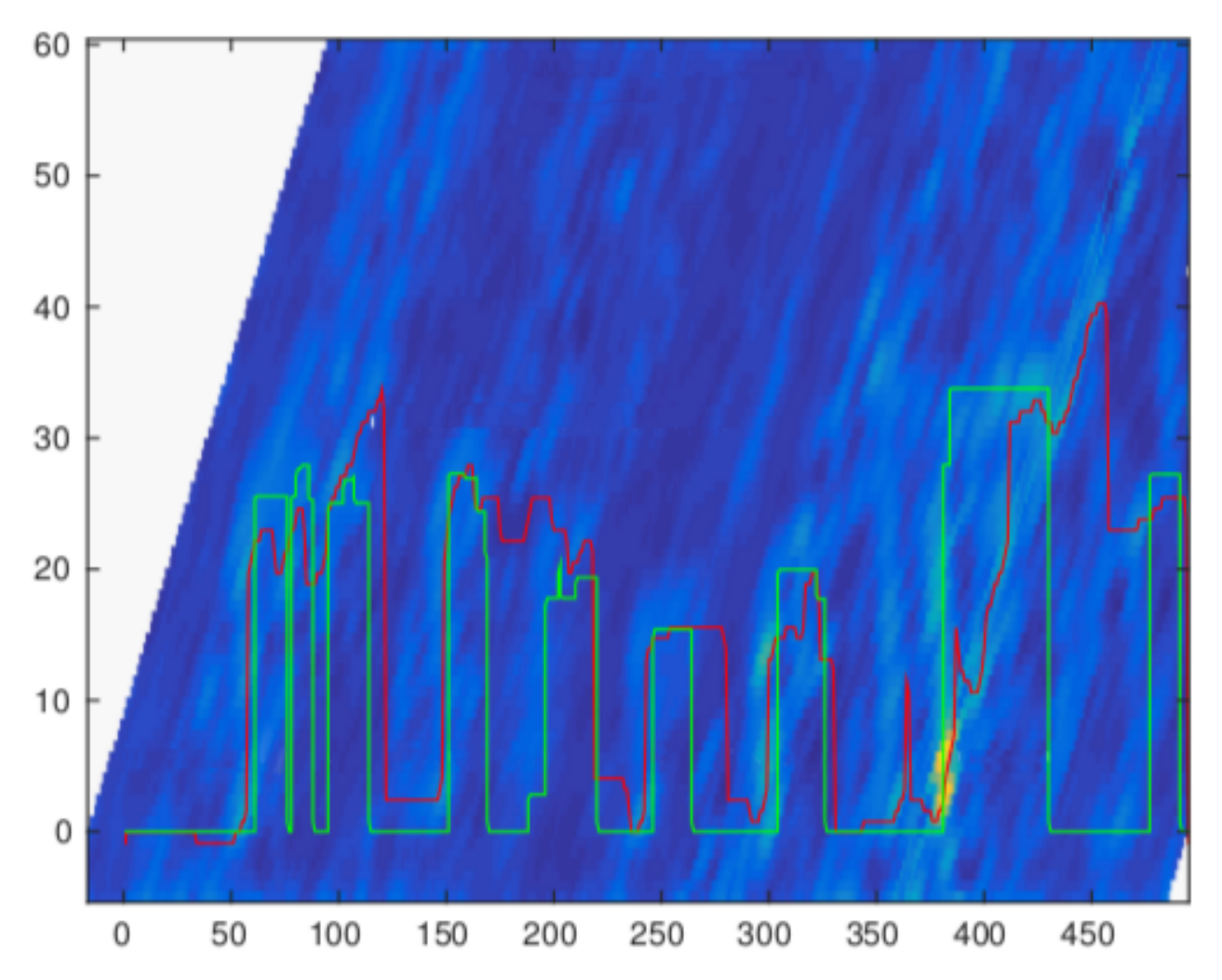}};
			
			\node[below=of img, node distance=0cm,xshift=0.cm, yshift=1.1cm, scale = .6
			,font=\color{black}] {y};
			
			\node[left=of img, node distance=0cm, rotate=0, scale = .6,
			anchor=center,yshift=0.45cm,xshift=1.7cm,font=\color{black}] {z};
			\end{tikzpicture}
		}
	\end{minipage}
	\begin{minipage}{\linewidth}
		\subfloat[]{
			\begin{tikzpicture}
			\node (img)  {\includegraphics[width=.44\linewidth]{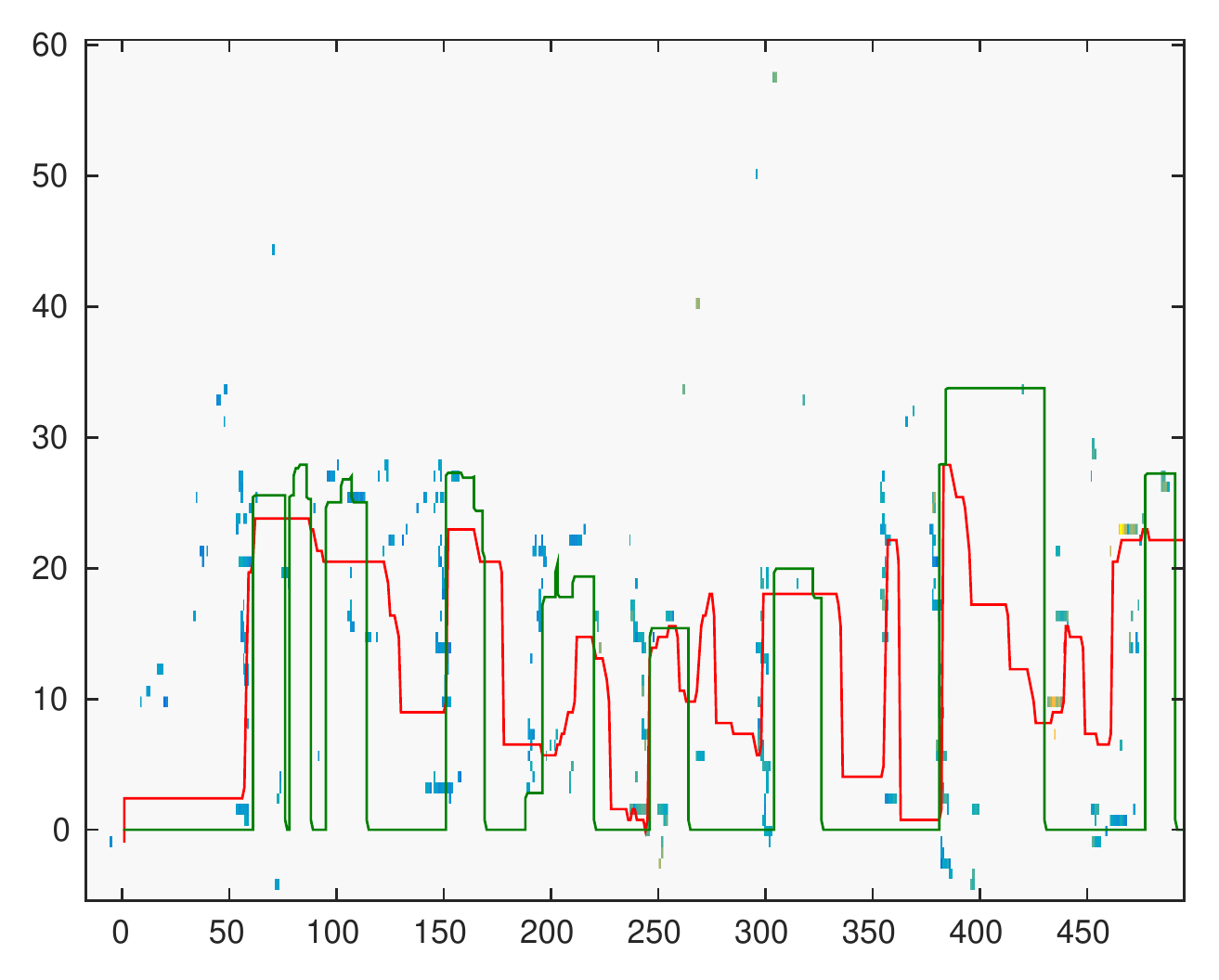}};
			
			\node[below=of img, node distance=0cm,xshift=0.cm, yshift=1.1cm, scale = .6
			,font=\color{black}] {y};
			
			\node[left=of img, node distance=0cm, rotate=0, scale = .6,
			anchor=center,yshift=0.45cm,xshift=1.7cm,font=\color{black}] {z};
			\end{tikzpicture}
		}\hfil
		\subfloat[]{
			\begin{tikzpicture}
			\node (img)  {\includegraphics[width=.44\linewidth]{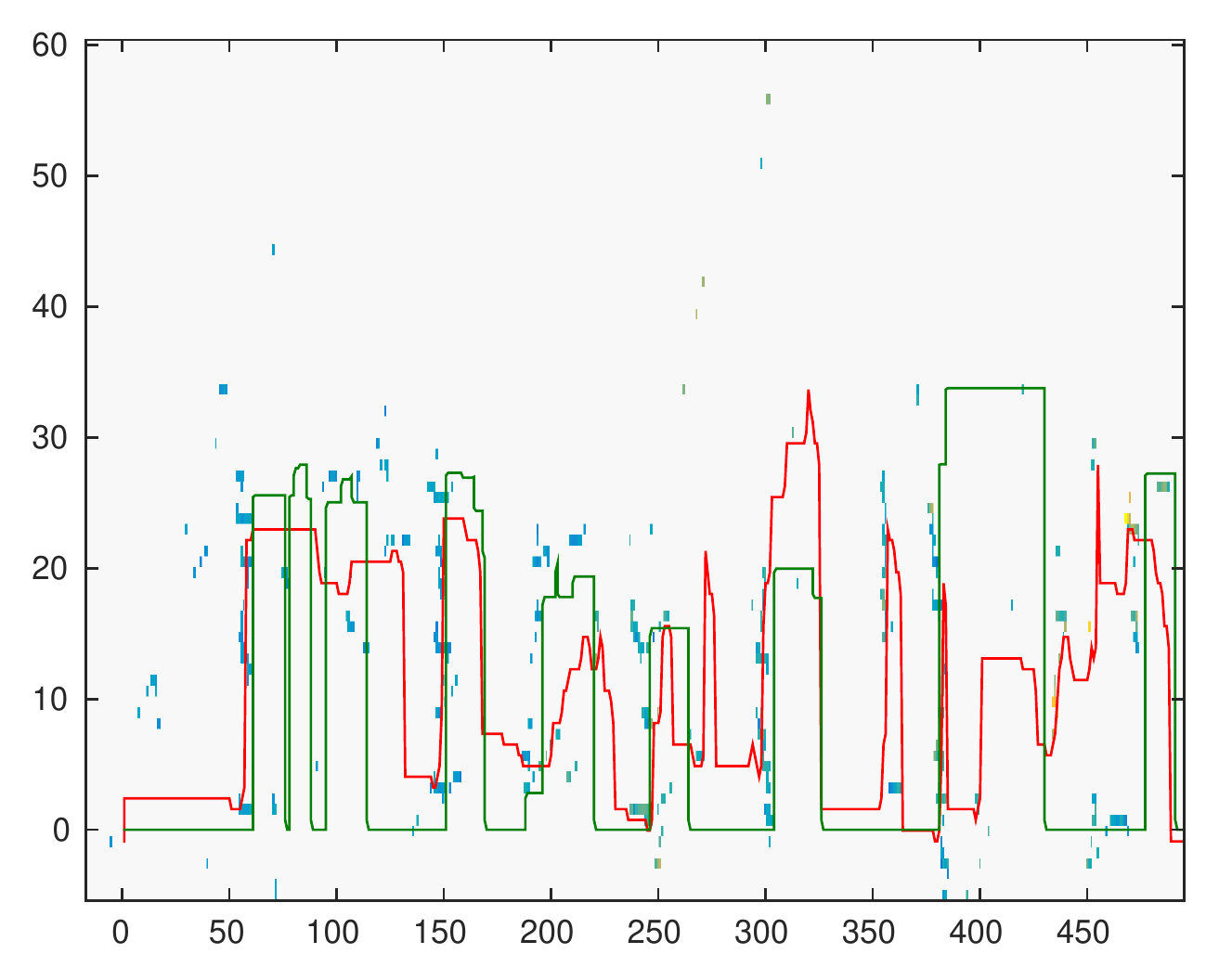}};
			
			\node[below=of img, node distance=0cm,xshift=0.cm, yshift=1.1cm, scale = .6
			,font=\color{black}] {y};
			
			\node[left=of img, node distance=0cm, rotate=0, scale = .6,
			anchor=center,yshift=0.45cm,xshift=1.7cm,font=\color{black}] {z};
			\end{tikzpicture}
		}
	\end{minipage}
	\begin{minipage}{\linewidth}
		\subfloat[]{
			\begin{tikzpicture}
			\node (img)  {\includegraphics[width=.44\linewidth]{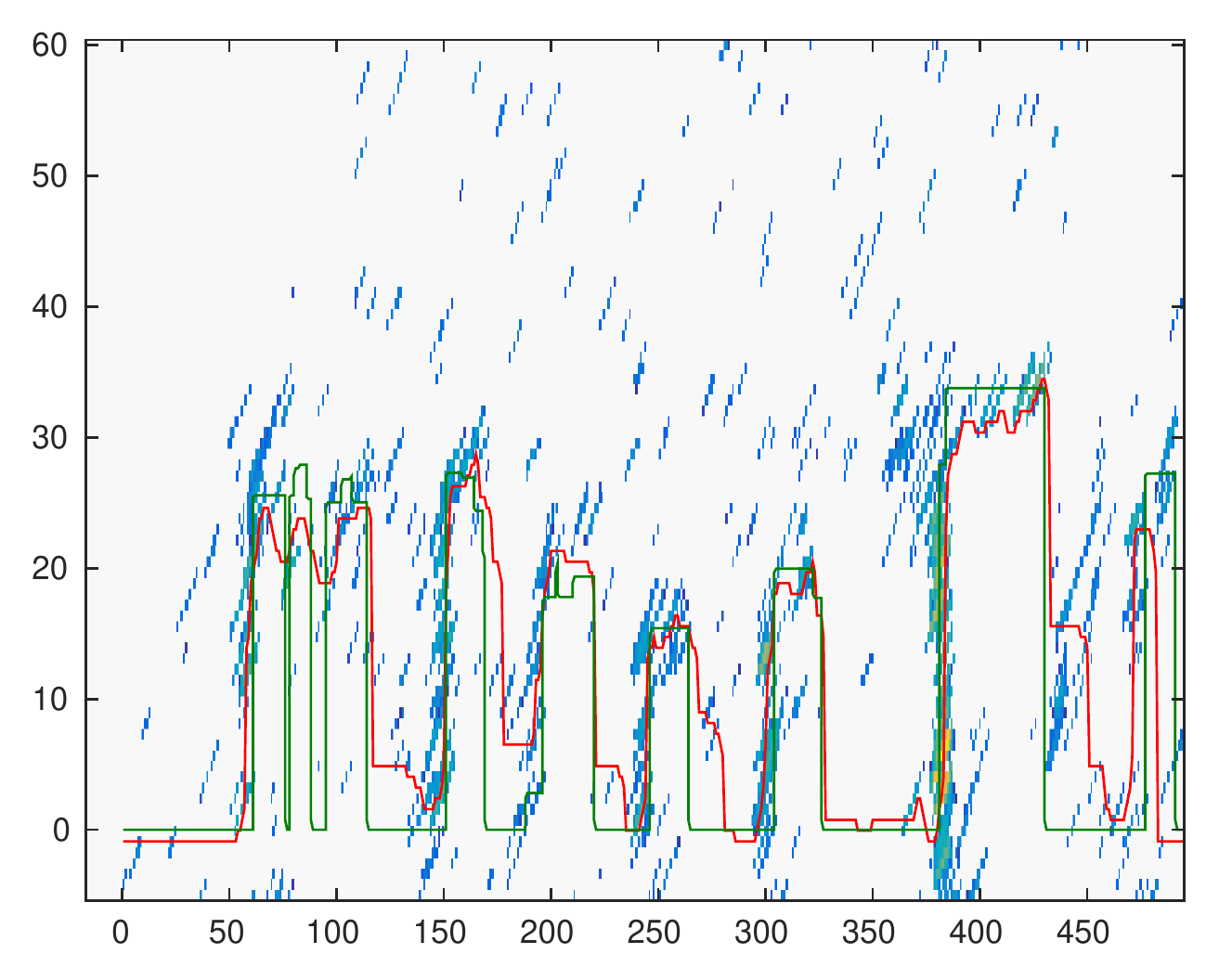}};
			
			\node[below=of img, node distance=0cm,xshift=0.cm, yshift=1.1cm, scale = .6
			,font=\color{black}] {y};
			
			\node[left=of img, node distance=0cm, rotate=0, scale = .6,
			anchor=center,yshift=0.45cm,xshift=1.7cm,font=\color{black}] {z};
			\end{tikzpicture}
		}\hfil
		\subfloat[]{
			\begin{tikzpicture}
			\node (img)  {\includegraphics[width=.44\linewidth]{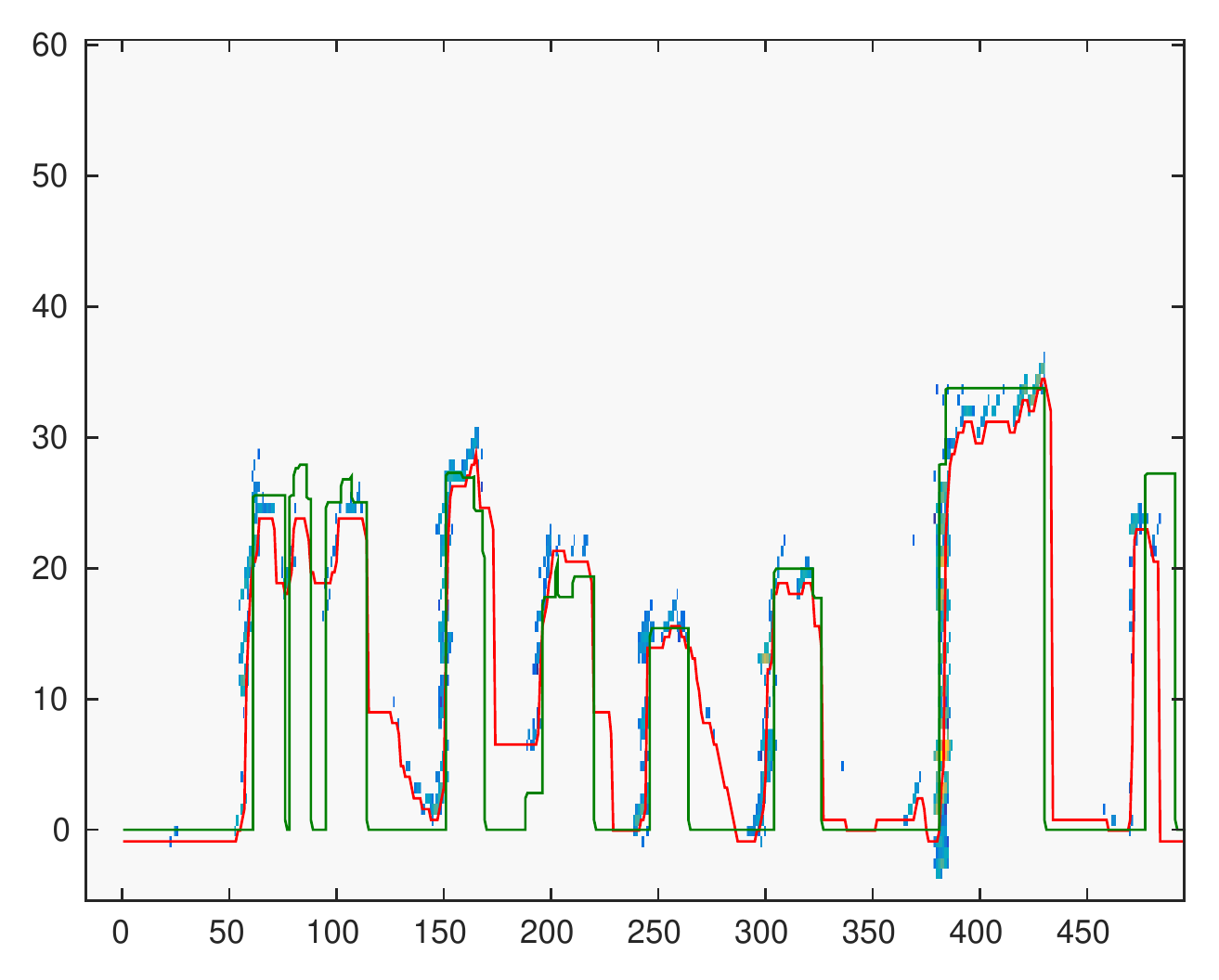}};
			
			\node[below=of img, node distance=0cm,xshift=0.cm, yshift=1.1cm, scale = .6
			,font=\color{black}] {y};
			
			\node[left=of img, node distance=0cm, rotate=0, scale = .6,
			anchor=center,yshift=0.45cm,xshift=1.7cm,font=\color{black}] {z};
			\end{tikzpicture}
		}
	\end{minipage}
	\caption{Urban surface estimation using graph-cut segmentation of the
		tomograms, as described in section 3. The estimated surface corresponds to the
		red profile. The ground truth for the given slice is shown in green. The
		tomograms are obtained using Capon beamforming (a), SPICE (b), MUSIC (c), WSF (d), the 3-D inversion
		approach (e) and REDRESS (f).}
	\label{fig:slice}
\end{figure}
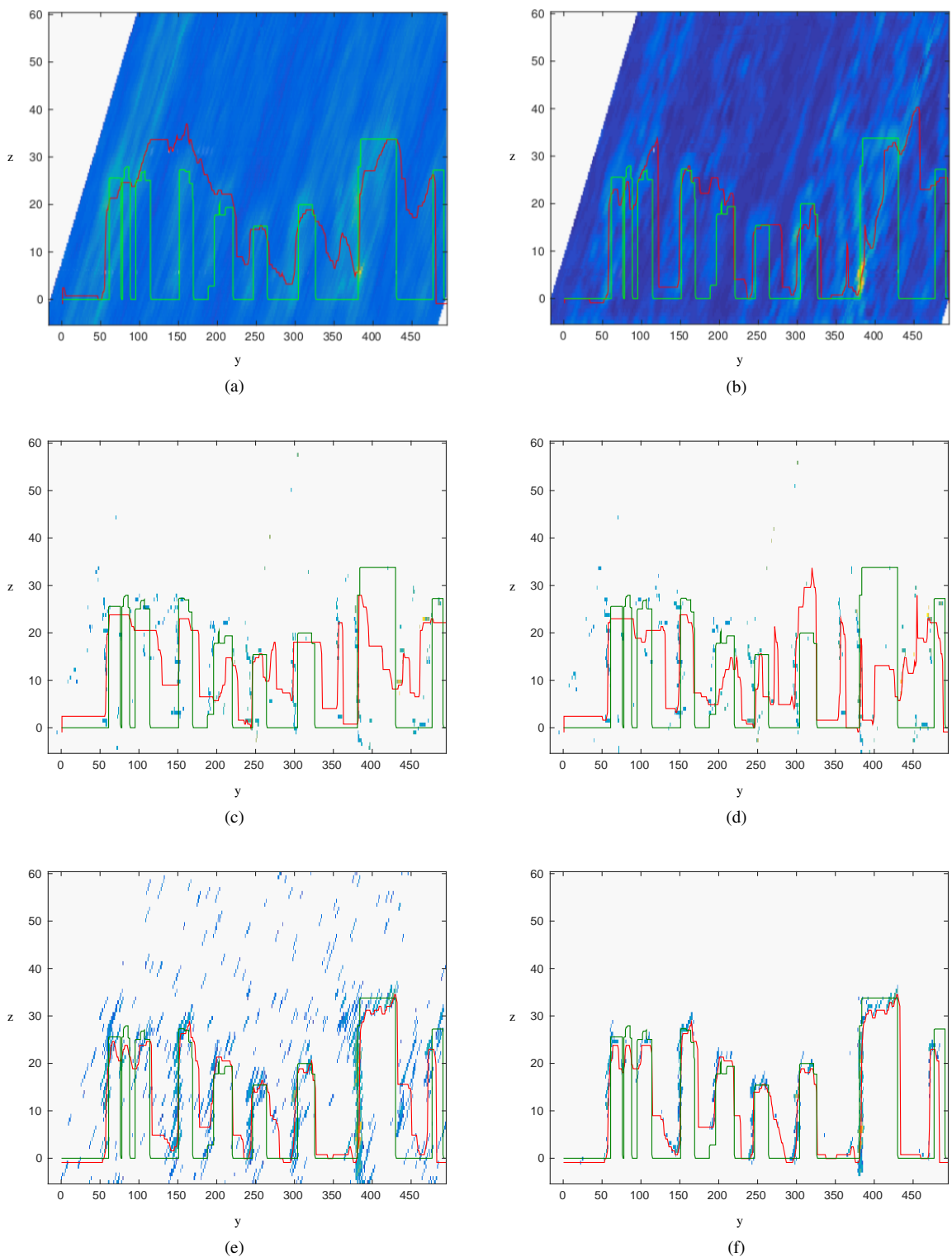

\begin{figure*}[!ht]
	\centering
	\begin{minipage}{.21\linewidth}
		\centering
		\subfloat[]{\includegraphics[trim={0 0 0 2cm},clip,width = .95\linewidth, height =
			6cm]{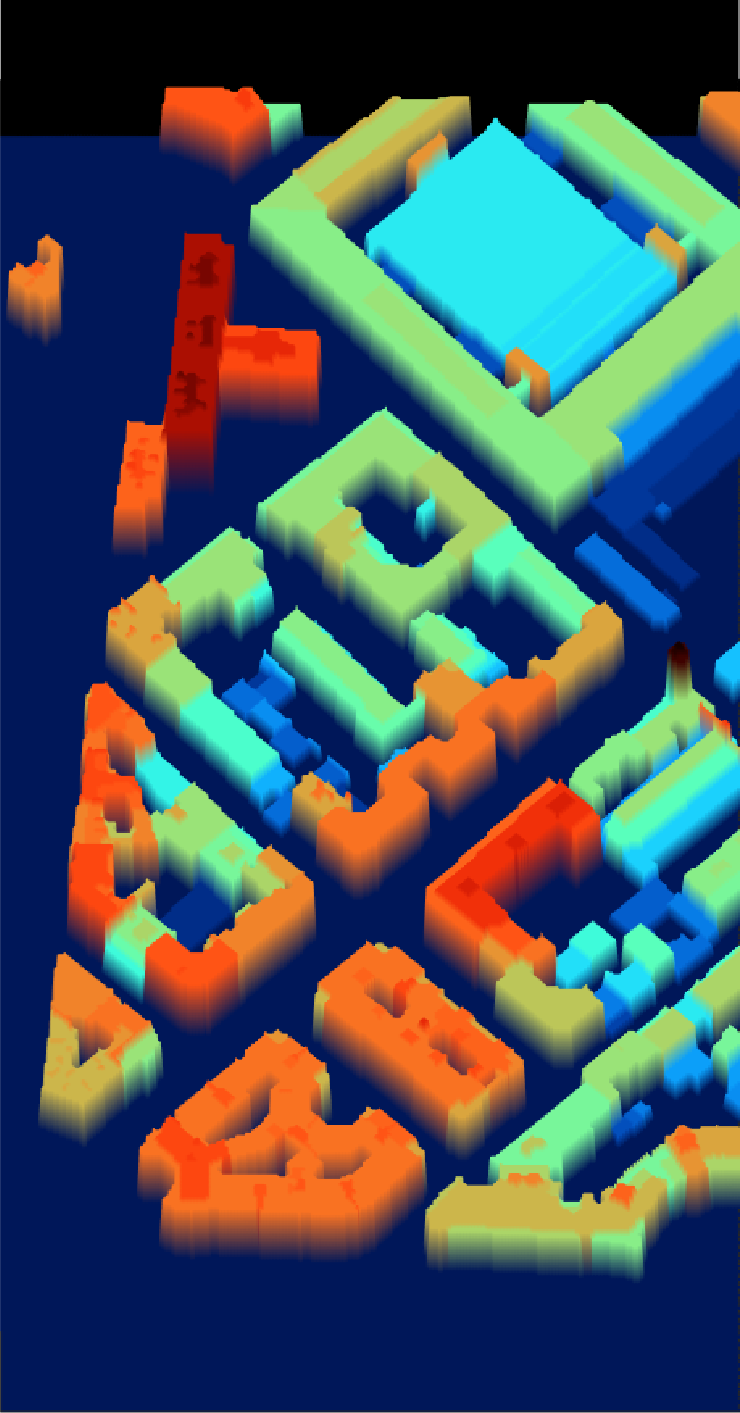}
		}\hfil
	\end{minipage} 
	\begin{minipage}{.21\linewidth}
		\centering
		\subfloat[]{\includegraphics[trim={0 0 0 2cm},clip,width = .95\linewidth, height =
			6cm]{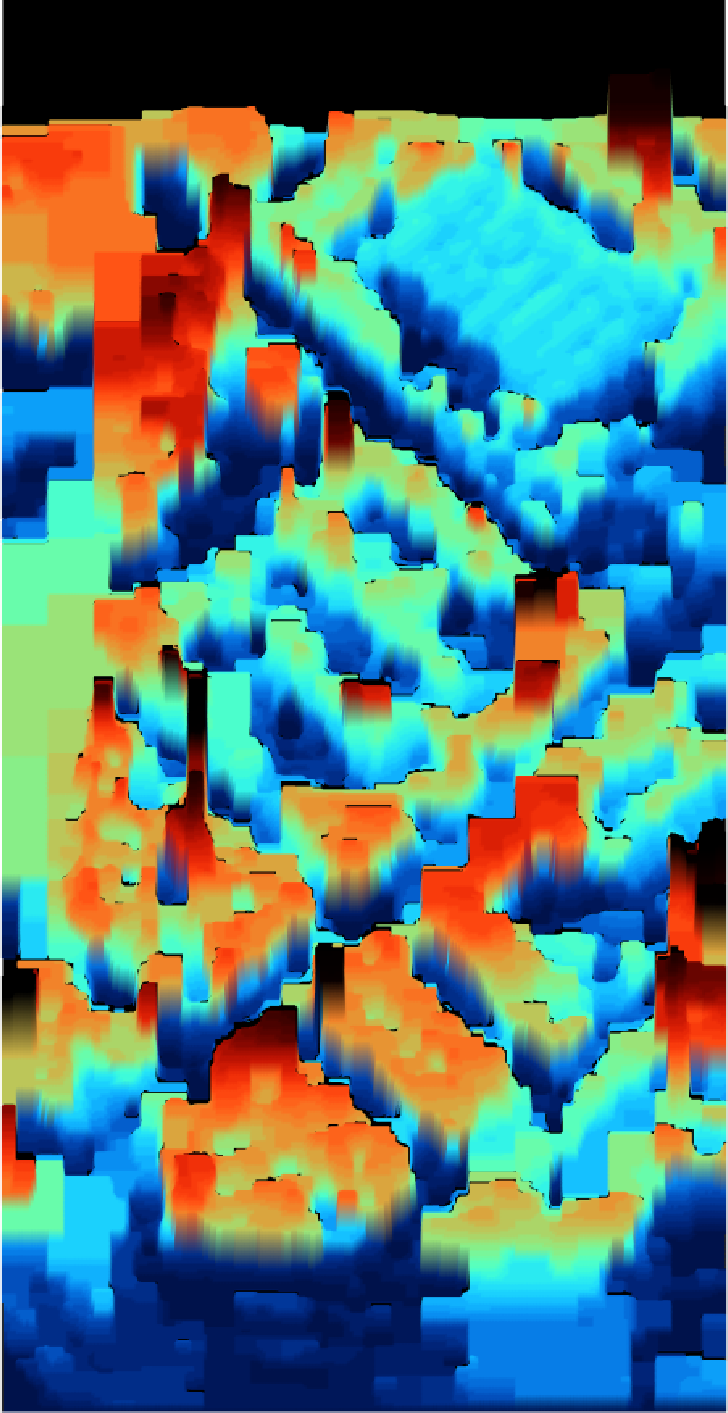}
		}\hfil
	\end{minipage} 
	\begin{minipage}{.21\linewidth}
		\centering
		\subfloat[]{\includegraphics[trim={0 0 0 2cm},clip,width = .95\linewidth, height =
			6cm]{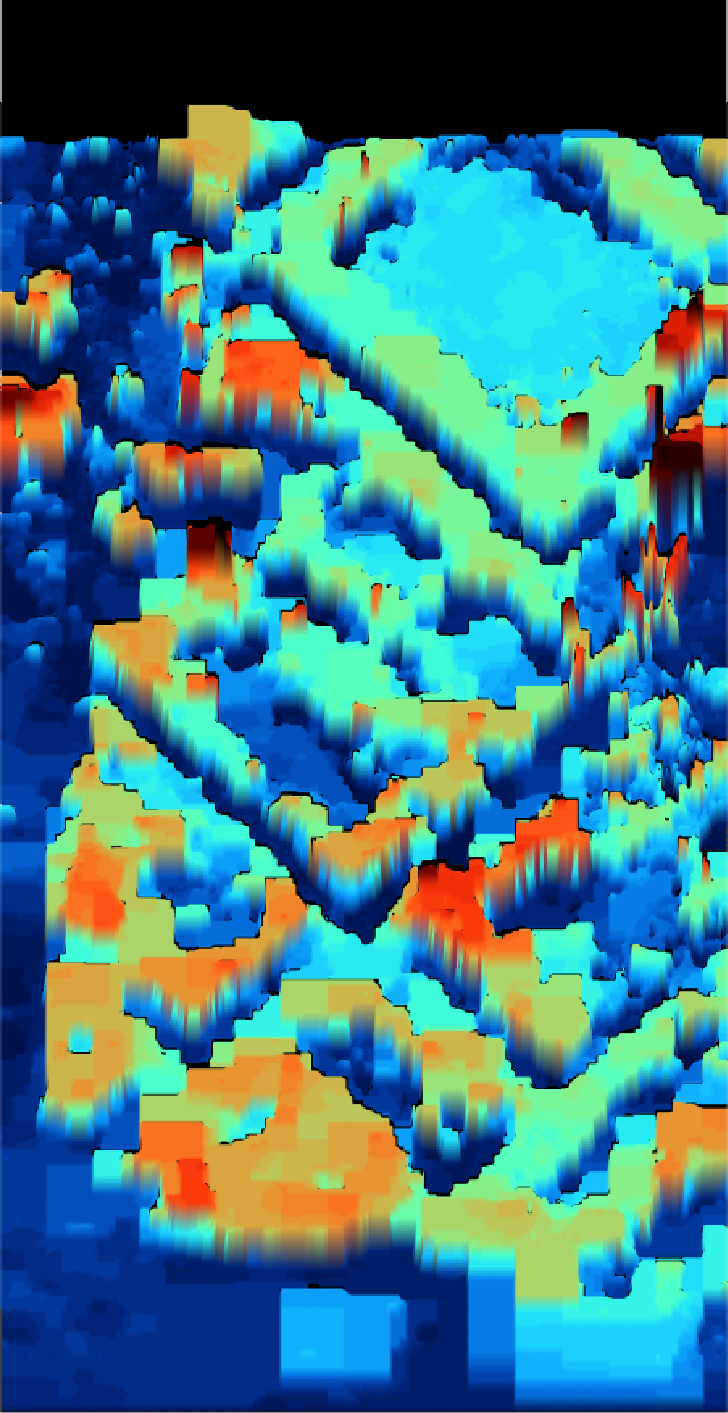}
		}\hfil
	\end{minipage} 
	\begin{minipage}{.21\linewidth}
		\centering
		\subfloat[]{\includegraphics[trim={0 0 0 2cm},clip,width = .95\linewidth, height =
			6cm]{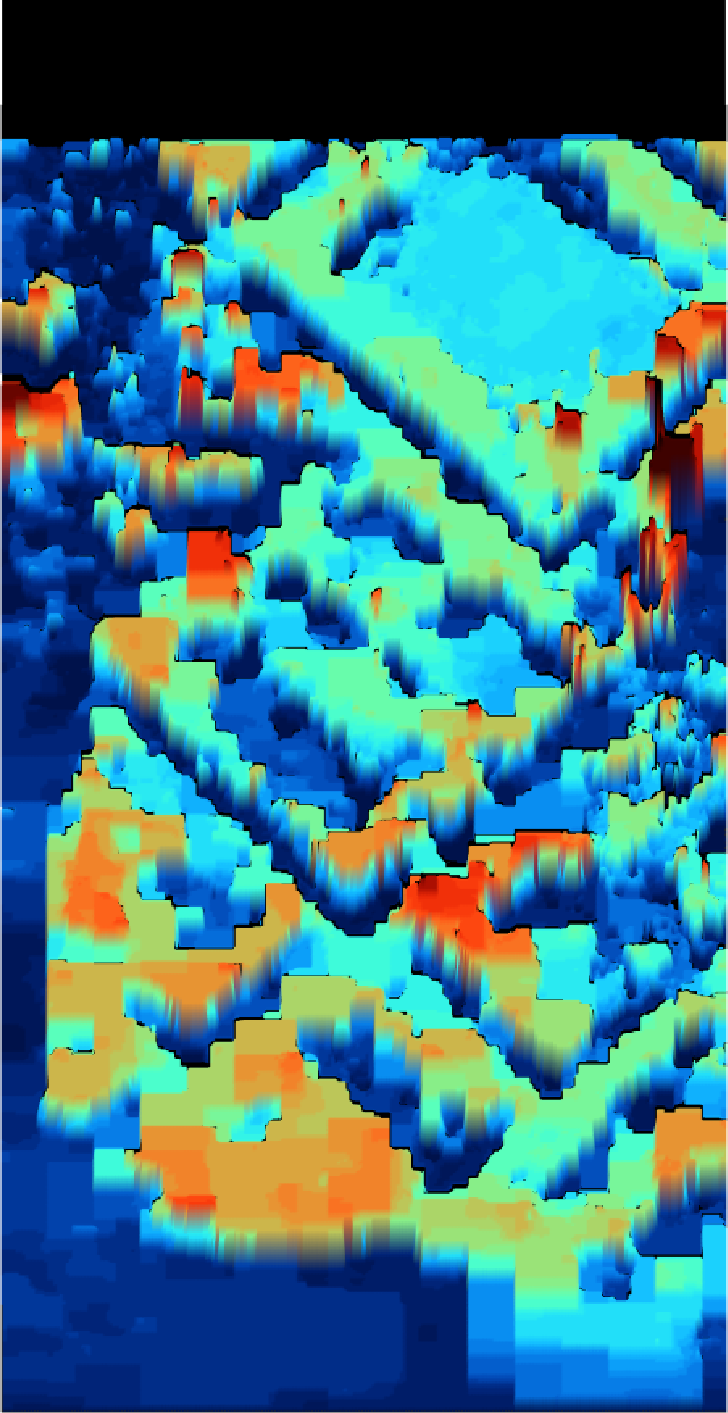}
		}\hfil
	\end{minipage}
	\begin{minipage}{.21\linewidth}
		\centering
		\subfloat[]{\includegraphics[trim={0 0 0 2cm},clip,width = .95\linewidth, height =
			6cm]{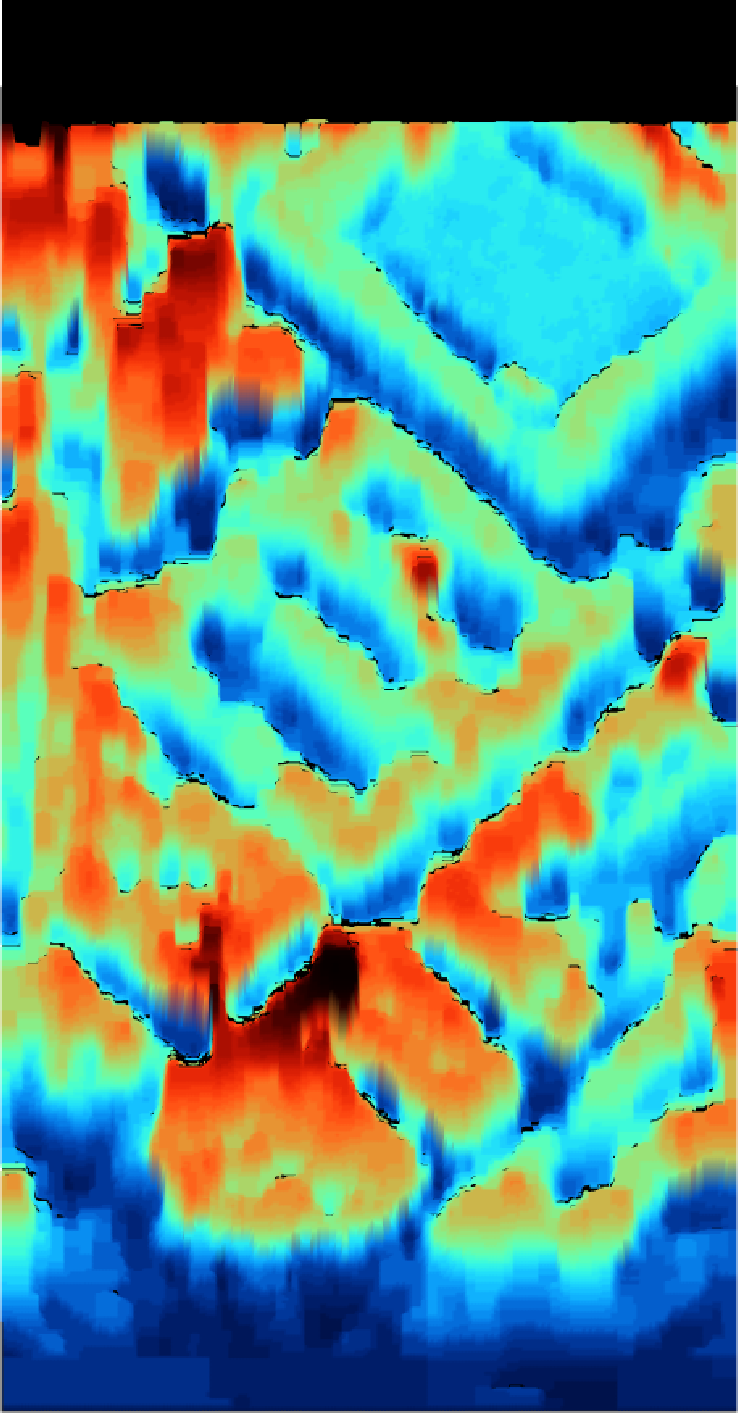}
		}\hfil
	\end{minipage} 
	\begin{minipage}{.21\linewidth}
		\centering
		\subfloat[]{\includegraphics[trim={0 0 0 2cm},clip,width = .95\linewidth, height =
			6cm]{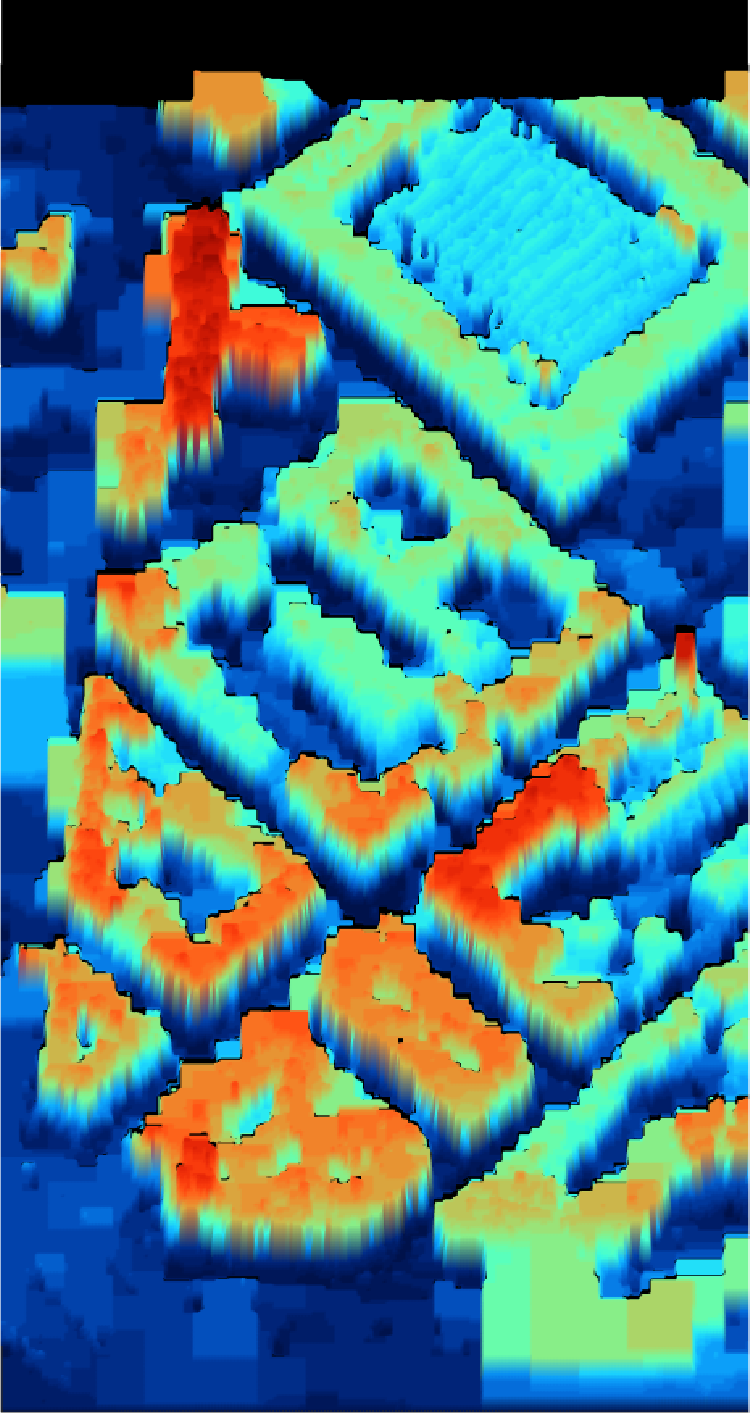}
		}\hfil
	\end{minipage} 
	\begin{minipage}{.21\linewidth}
		\centering
		\subfloat[]{\includegraphics[trim={0 0 0 2cm},clip,width = .95\linewidth, height =
			6cm]{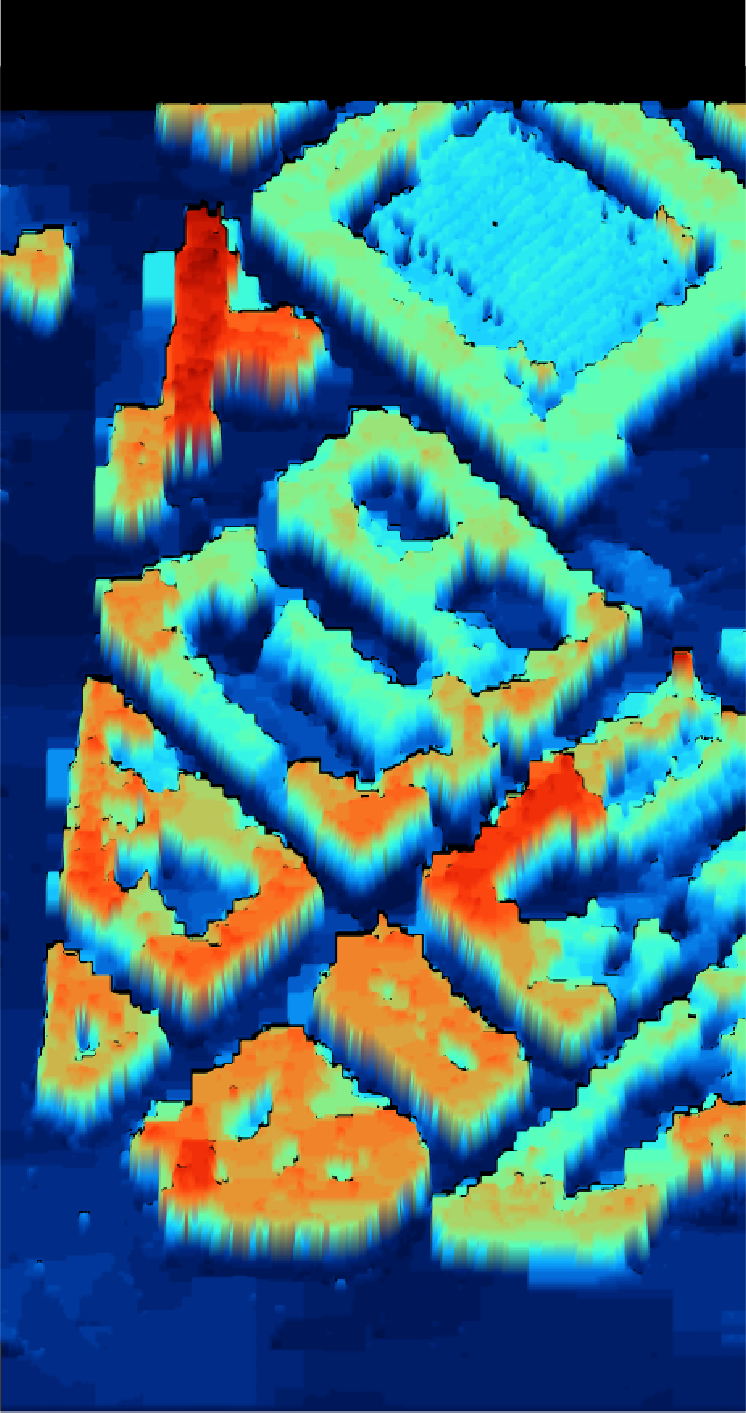}
		}\hfil
	\end{minipage} 
		\caption{Ground truth height (a), scene surface estimation using SPICE (b), MUSIC
			(c), WSF (d), Capon beamforming (e), the 3-D inversion (f) and REDRESS (g). For each results, the image
			shows the surface colored according to its
			height.}
		\label{fig:3D_view}
	\end{figure*}

\begin{figure*}[!ht]
		\centering
		\begin{minipage}{.21\linewidth}
			\centering
			\subfloat[]{\includegraphics[trim={0 0 0 2cm},clip,width = .95\linewidth, height =
				6cm]{bdtopo_view_b.png}
			}\hfil
		\end{minipage} 
		\begin{minipage}{.21\linewidth}
			\centering
			\subfloat[]{\includegraphics[trim={0 0 0 2cm},clip,width = .95\linewidth, height =
				6cm]{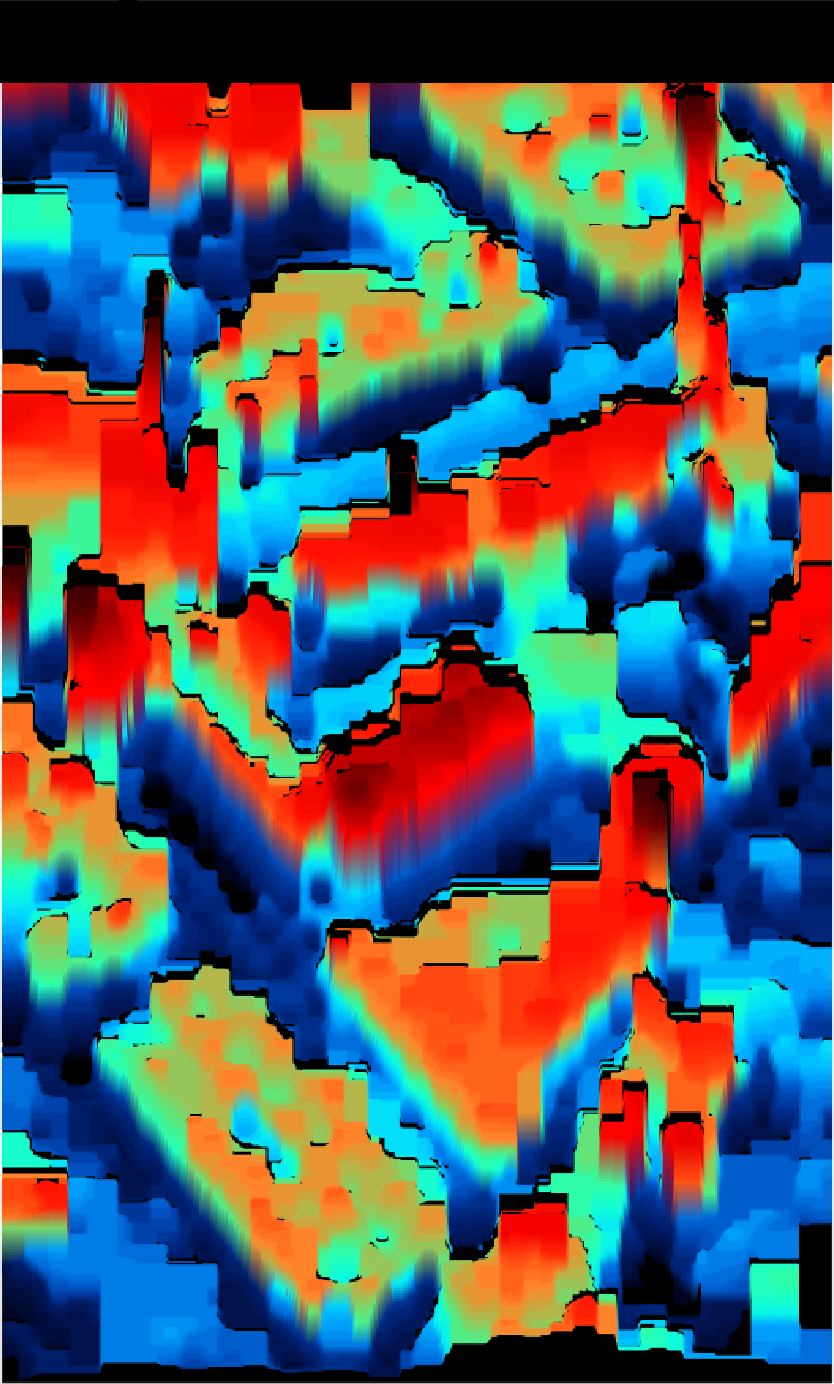}
			}\hfil
		\end{minipage} 
		\begin{minipage}{.21\linewidth}
			\centering
			\subfloat[]{\includegraphics[trim={0 0 0 2cm},clip,width = .95\linewidth, height =
				6cm]{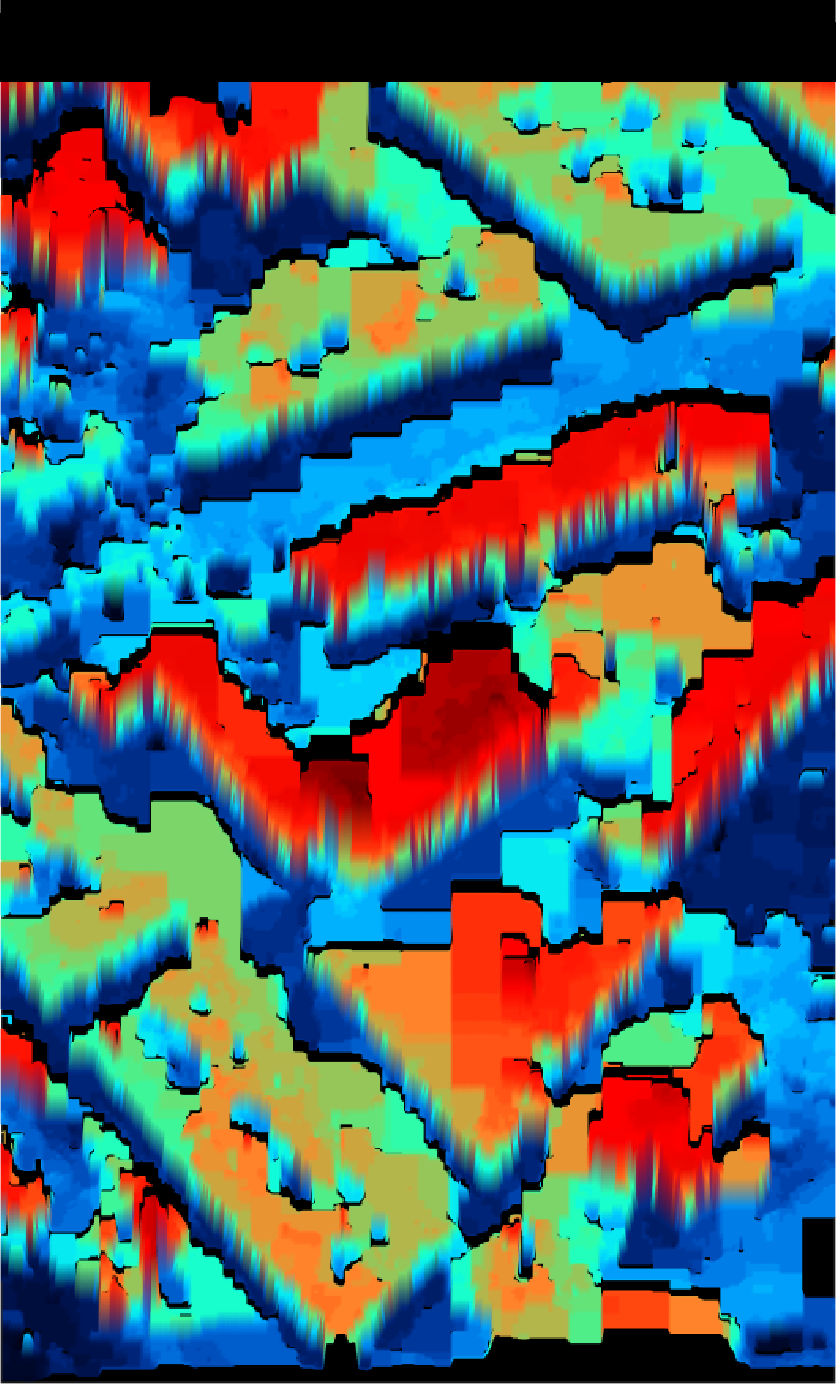}
			}\hfil
		\end{minipage} 
		\begin{minipage}{.21\linewidth}
			\centering
			\subfloat[]{\includegraphics[trim={0 0 0 2cm},clip,width = .95\linewidth, height =
				6cm]{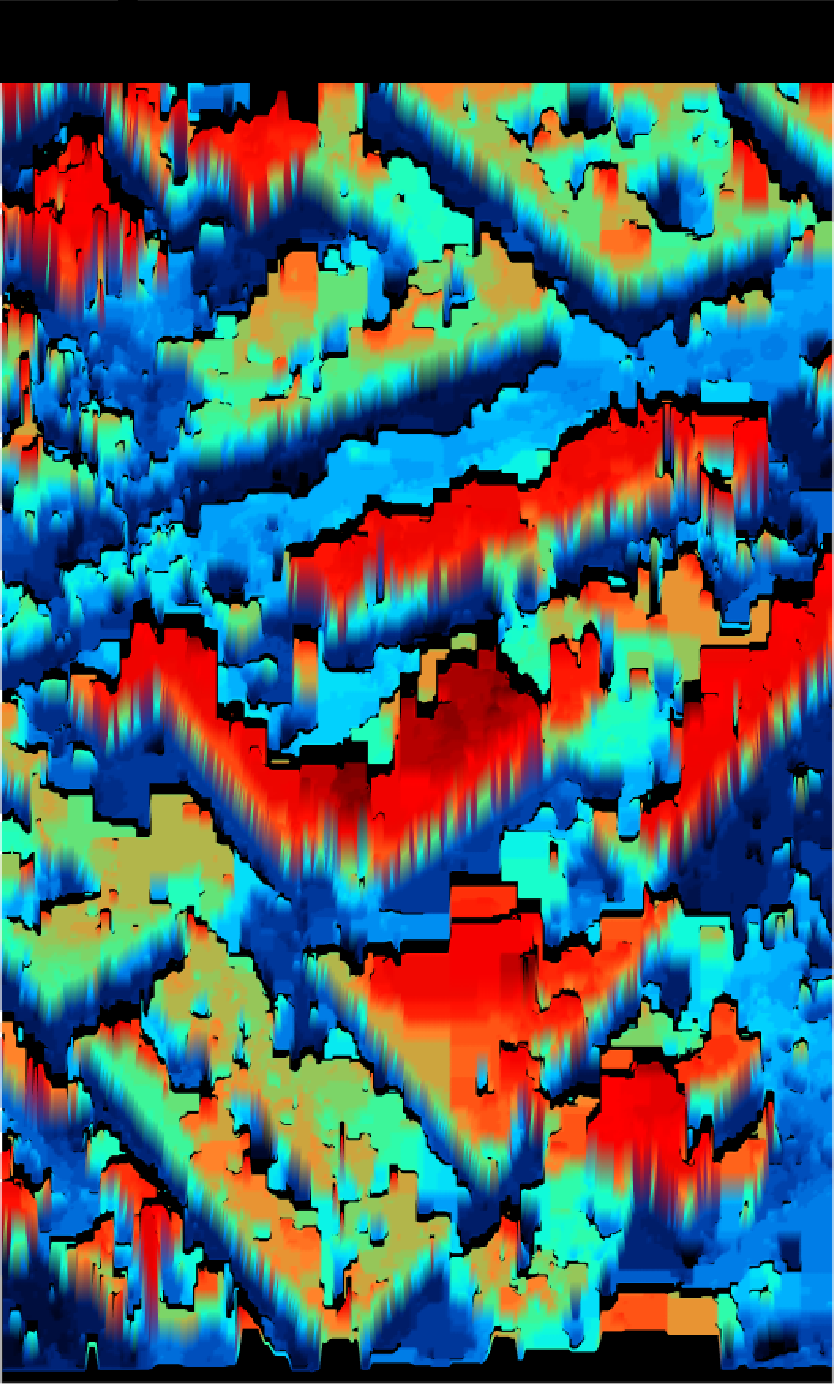}
			}\hfil
		\end{minipage}
		\begin{minipage}{.21\linewidth}
			\centering
			\subfloat[]{\includegraphics[trim={0 0 0 2cm},clip,width = .95\linewidth, height =
				6cm]{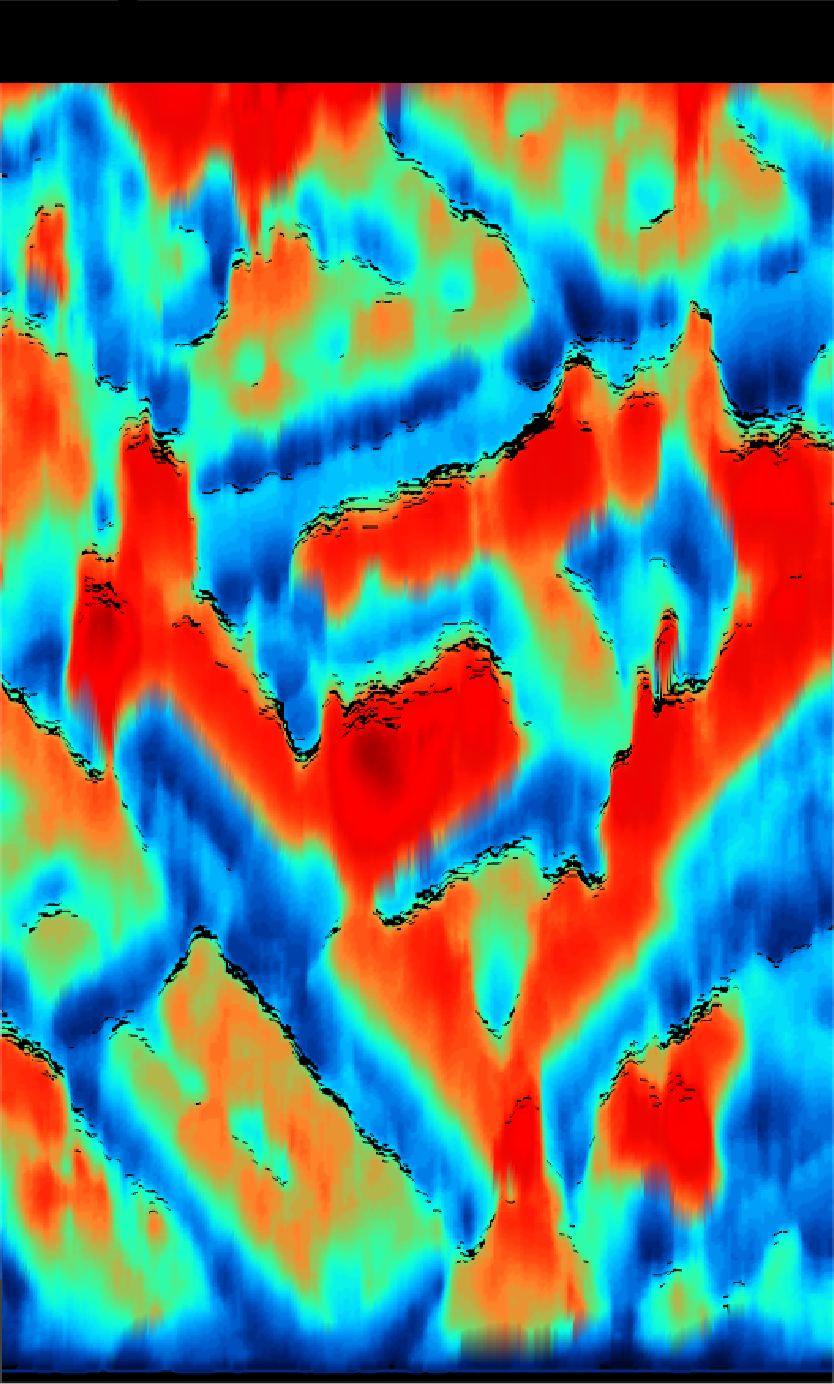}
			}\hfil
		\end{minipage} 
		\begin{minipage}{.21\linewidth}
			\centering
			\subfloat[]{\includegraphics[trim={0 0 0 2cm},clip,width = .95\linewidth, height =
				6cm]{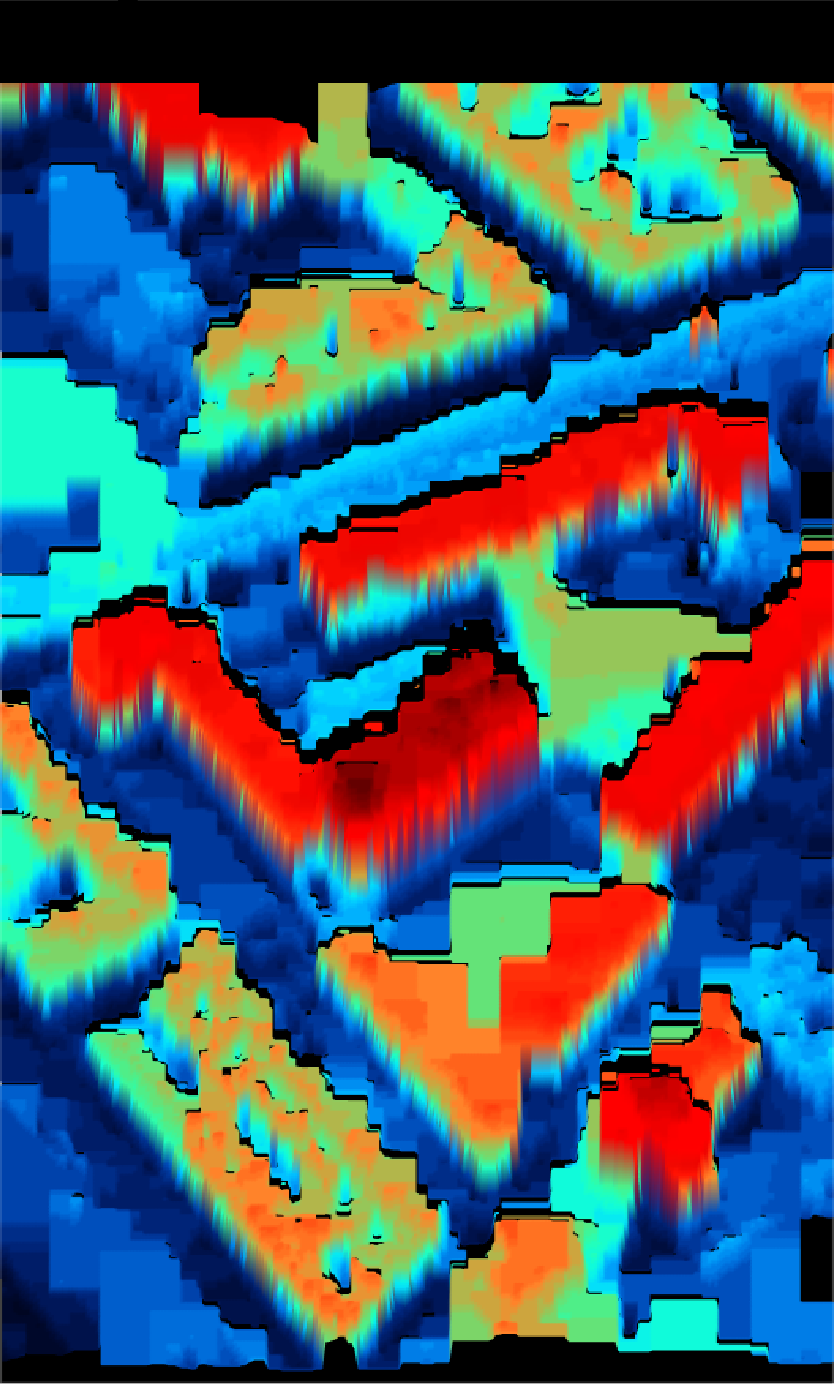}
			}\hfil
		\end{minipage} 
		\begin{minipage}{.21\linewidth}
			\centering
			\subfloat[]{\includegraphics[trim={0 0 0 2cm},clip,width = .95\linewidth, height =
				6cm]{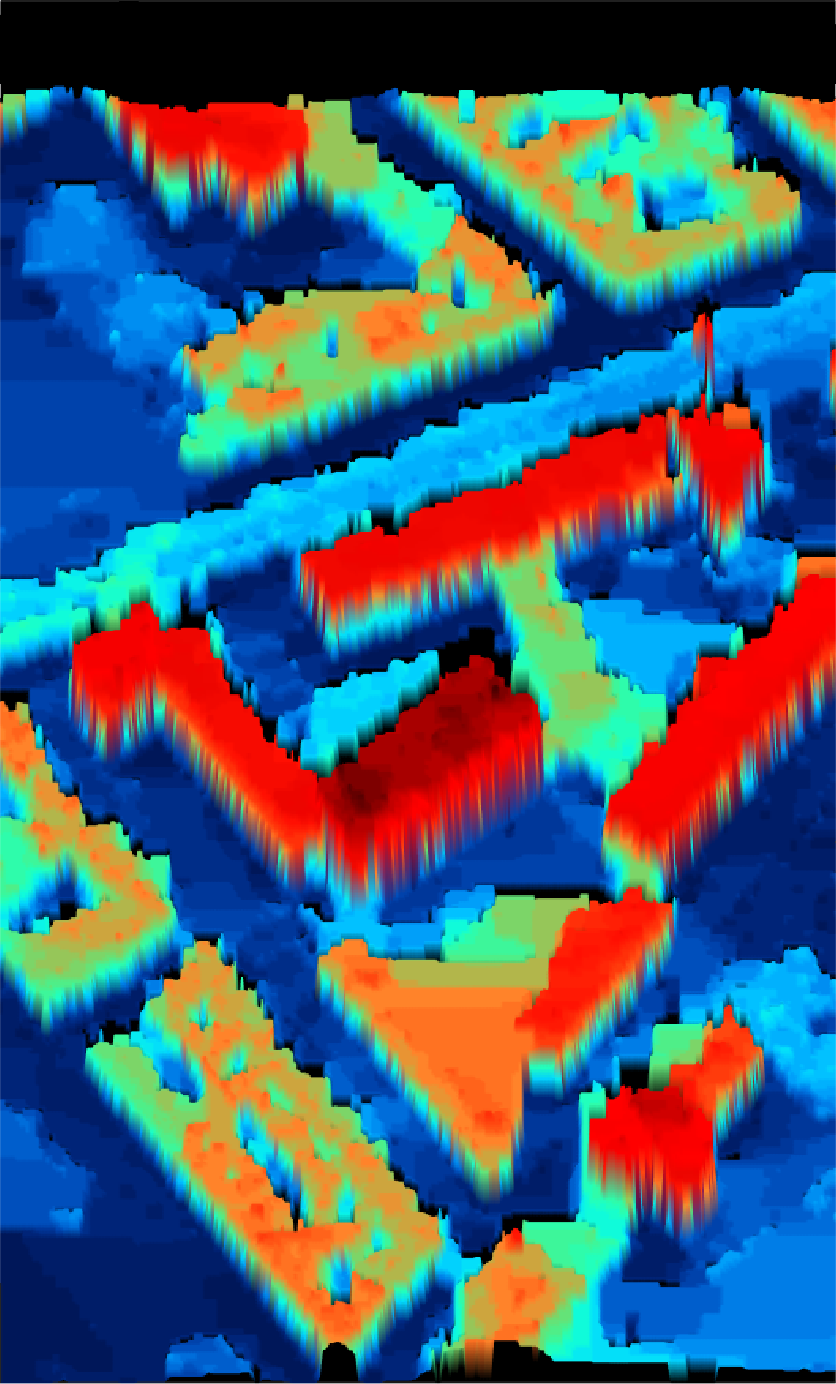}
			}\hfil
		\end{minipage} 
		\caption{Ground truth height (a), scene surface estimation using SPICE (b), MUSIC
			(c), WSF (d), Capon beamforming (e), the 3-D inversion (f) and REDRESS (g). For each results, the image
			shows the surface colored according to its
			height.}
		\label{fig:3D_view_b}
	\end{figure*}

The results are shown in Fig. \ref{fig:slice} for the
reconstructions of the slice, and
in Fig. \ref{fig:3D_view} and Fig. \ref{fig:3D_view_b} for the reconstruction of the whole
scenes. With the first experiment 
the behavior of each estimator and the resulting surface can
be observed in greater details. In the reconstructions, the areas where the surface is occluding itself are detected as shadow areas and removed. The resulting gaps
introduced are filled according to the height of the first point
outside it.

\begin{figure}[!h]
	\begin{minipage}{\linewidth}
		\centering
		\subfloat[]{
			\begin{tikzpicture}
			\node (img)  {\includegraphics[width=.4\linewidth]{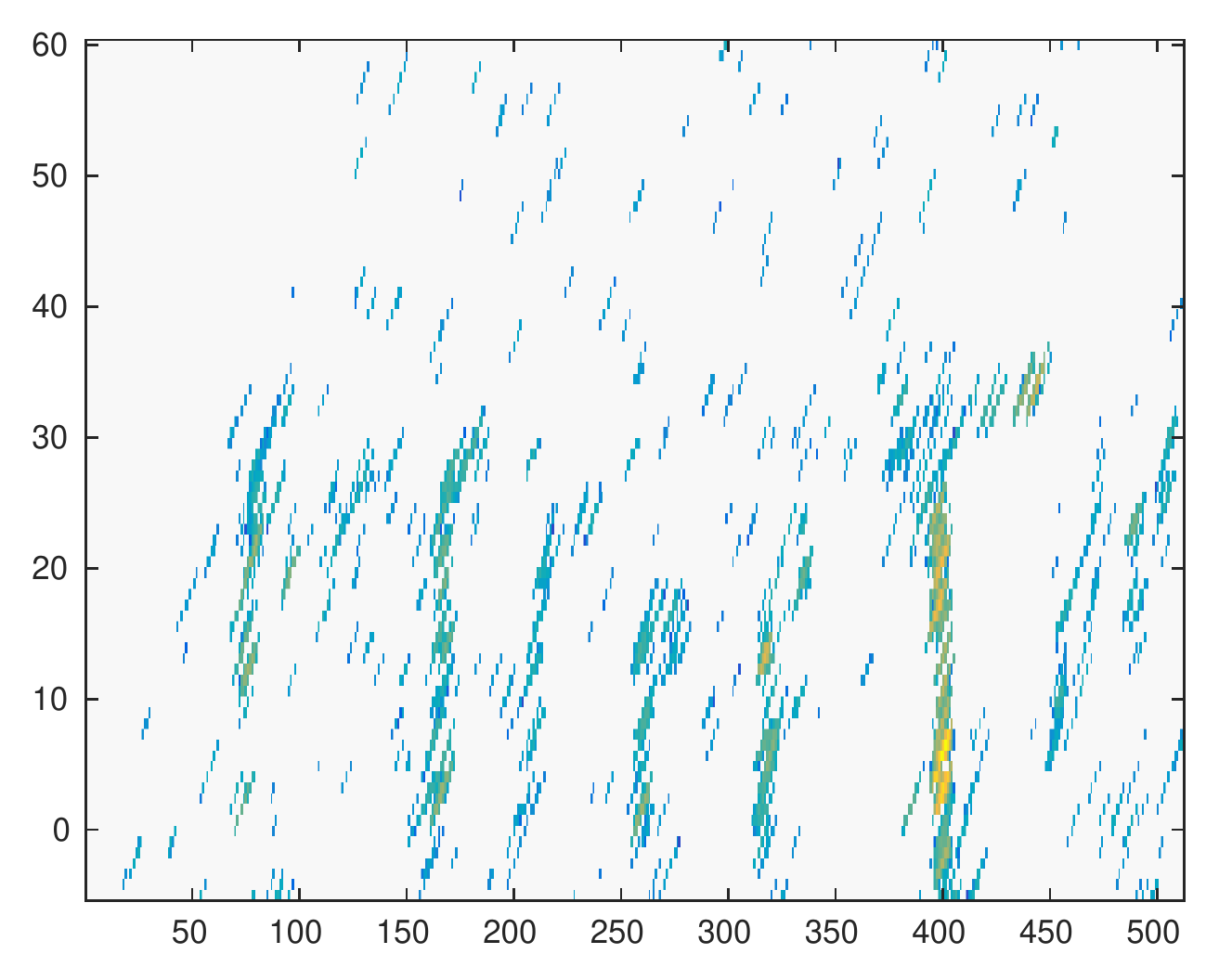}};
			
			\node[below=of img, node distance=0cm,xshift=0.cm, yshift=1.1cm, scale = .6
			,font=\color{black}] {y};
			
			\node[left=of img, node distance=0cm, rotate=0, scale = .6,
			anchor=center,yshift=0.45cm,xshift=1.7cm,font=\color{black}] {z};
			\end{tikzpicture}
			\begin{tikzpicture}
			\node (img)  {\includegraphics[width=.4\linewidth]{iter1_prof}};
			
			\node[below=of img, node distance=0cm,xshift=0.cm, yshift=1.1cm, scale = .6
			,font=\color{black}] {y};
			
			\node[left=of img, node distance=0cm, rotate=0, scale = .6,
			anchor=center,yshift=0.45cm,xshift=1.7cm,font=\color{black}] {z};
			\end{tikzpicture}}\hfil
	\end{minipage}
	\begin{minipage}{\linewidth}
		\centering
		\subfloat[]{
			\begin{tikzpicture}
			\node (img)  {\includegraphics[width=.4\linewidth]{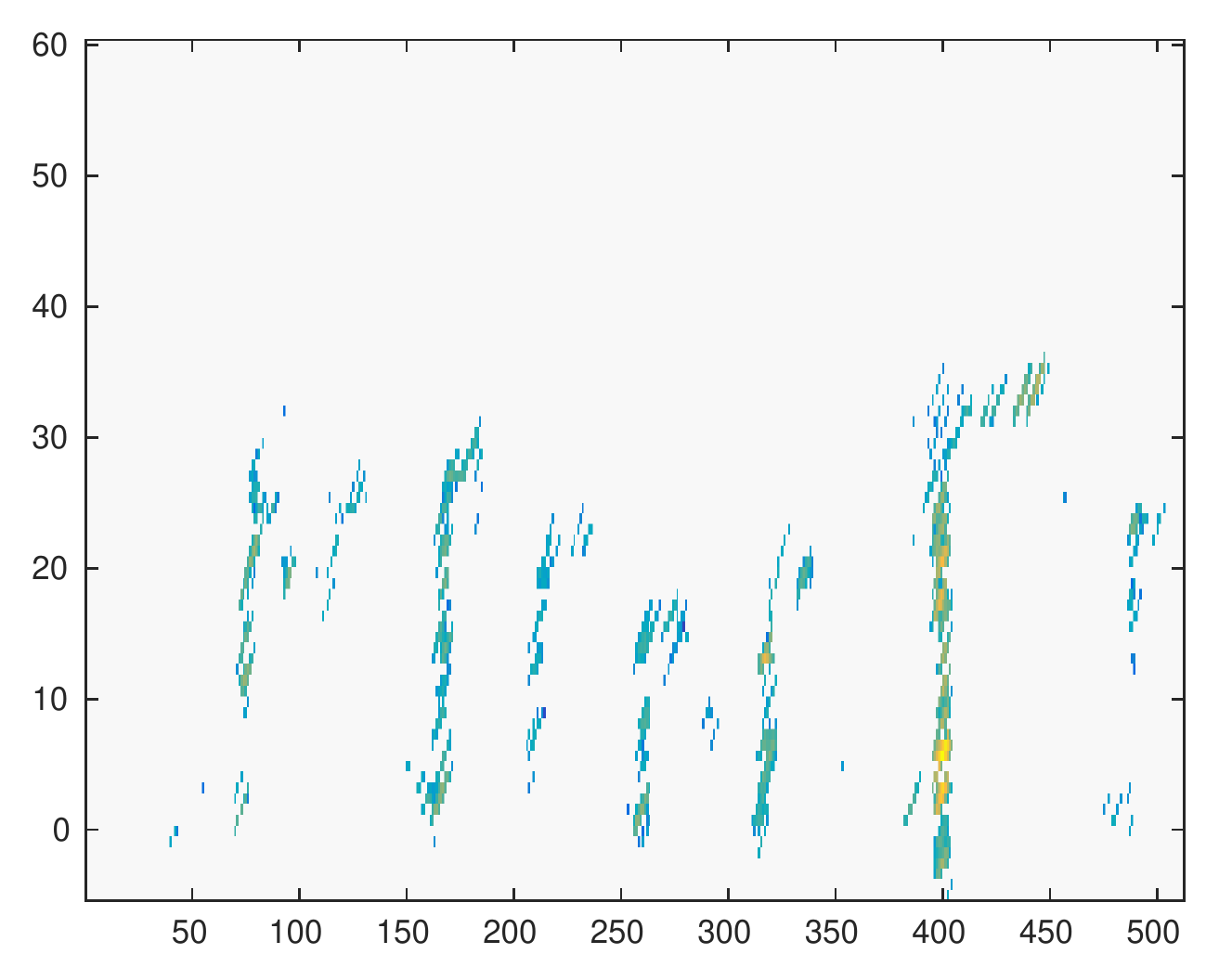}};
			
			\node[below=of img, node distance=0cm,xshift=0.cm, yshift=1.1cm, scale = .6
			,font=\color{black}] {y};
			
			\node[left=of img, node distance=0cm, rotate=0, scale = .6,
			anchor=center,yshift=0.45cm,xshift=1.7cm,font=\color{black}] {z};
			\end{tikzpicture}
			\begin{tikzpicture}
			\node (img)  {\includegraphics[width=.4\linewidth]{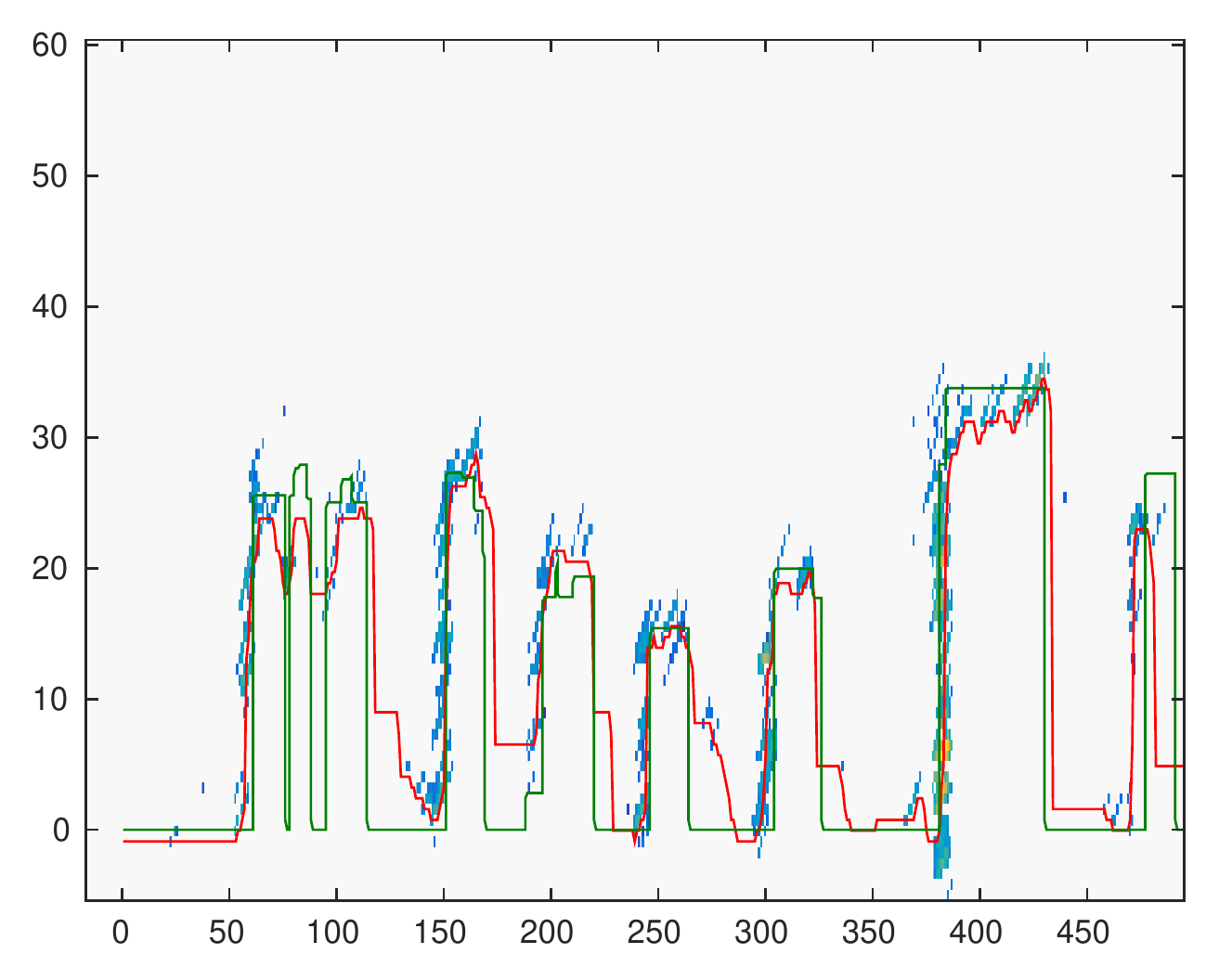}};
			
			\node[below=of img, node distance=0cm,xshift=0.cm, yshift=1.1cm, scale = .6
			,font=\color{black}] {y};
			
			\node[left=of img, node distance=0cm, rotate=0, scale = .6,
			anchor=center,yshift=0.45cm,xshift=1.7cm,font=\color{black}] {z};
			\end{tikzpicture}}\hfil
	\end{minipage}
	\begin{minipage}{\linewidth}
		\centering
		\subfloat[]{
			\begin{tikzpicture}
			\node (img)  {\includegraphics[width=.4\linewidth]{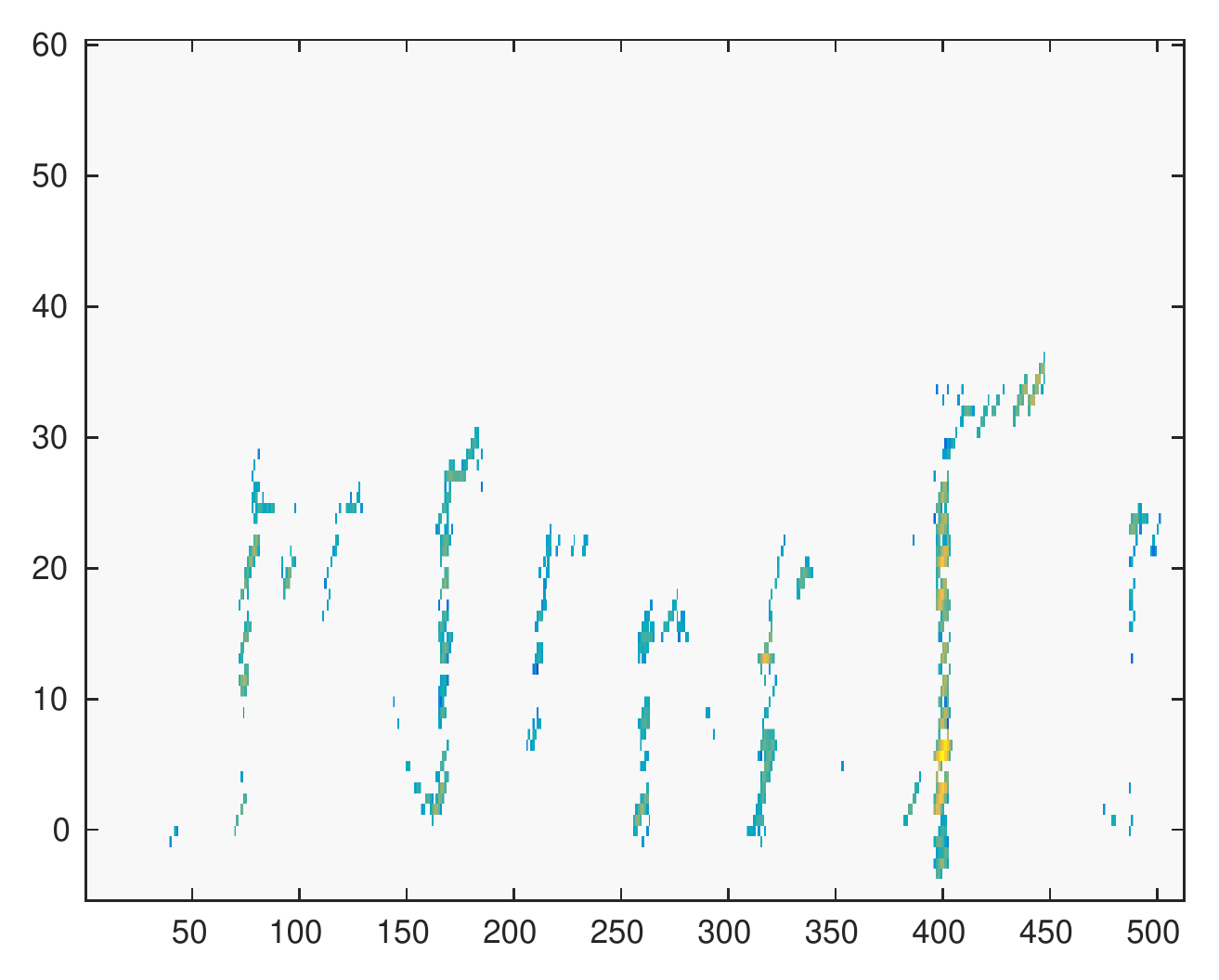}};
			
			\node[below=of img, node distance=0cm,xshift=0.cm, yshift=1.1cm, scale = .6
			,font=\color{black}] {y};
			
			\node[left=of img, node distance=0cm, rotate=0, scale = .6,
			anchor=center,yshift=0.45cm,xshift=1.7cm,font=\color{black}] {z};
			\end{tikzpicture}
			\begin{tikzpicture}
			\node (img)  {\includegraphics[width=.4\linewidth]{iter5_prof}};
			
			\node[below=of img, node distance=0cm,xshift=0.cm, yshift=1.1cm, scale = .6
			,font=\color{black}] {y};
			
			\node[left=of img, node distance=0cm, rotate=0, scale = .6,
			anchor=center,yshift=0.45cm,xshift=1.7cm,font=\color{black}] {z};
			\end{tikzpicture}}\hfil
	\end{minipage}
	\caption{Three different iteration steps from the alternate reconstruction
		algorithm. On the left column, the estimated
		reflectivities are shown for the profile
		presented in Fig \ref{fig:data}. On the right, the estimated surface (red) and
		the ground truth (green) are superimposed in addition to the estimated
		reflectivities.
		Rows (a), (b) and (c) correspond respectively to the first,
		third and fifth iterations (last one).}
	\label{fig:iter}
\end{figure}

A second experiment presents the evolution of the reconstructed slice as the REDRESS algorithm iterates \textit{cf.} Fig. \ref{fig:iter}. It can be observed that the
distribution of reflectivities becomes much sharper after a
few iterations.

The third experiment illustrates the role of
the 3-D smoothing for both scenes. The surface is shown as seen from the
sensor point of view. Since some
tomographic estimators provide an estimate of the
reflectivities, those reflectivities can be plotted to
illustrate the distribution of scatterers on the reconstructed
surfaces.

To estimate  the covariance matrix at each point, we used a
$7\times 7$ Gaussian filter.
For MUSIC and WSF, the number of scatterers is set constant and equal to 2 to
avoid selecting too many outliers while allowing multiple scatterers within each
radar resolution cell. For these two estimators, the reflectivity is estimated by
mean square minimization, to keep a physical interpretation of the tomograms. As
the scene is very heterogeneous with a lot of layover, this step introduces some
undesired mixing of the information in the image. The surfaces estimated from
tomographic reconstructions using spectral analysis techniques
present noticeable 
artifacts in the dense areas. Some structures are too
extended, partially filling
streets or the building atrium. Meanwhile, the averaging step makes the tomographic
estimation smoother in homogenous areas for the fully sparse approaches MUSIC
and WSF. 

The CS technique performed on all the data set presents results that seem visually the closest to the
ground truth. Many
details can be observed in this reconstruction: most of the rooftops and visible
streets are well segmented and the buildings atrium are also retrieved. 

For all the previous estimators, the TV minimization produces some building elongation
resulting in phantom structure in low intensity signal area. This can be seen
for instance in the bottom right part for Fig. \ref{fig:3D_view} or around the
position 450 for Fig. \ref{fig:slice}.

Finally, the reconstructions obtained using the REDRESS algorithm present the closest results to the ground truth with far less TV artifacts in dark areas. \revb{
However, even with the alternated approach, some gaps between close and bright structures are filled in the reconstruction, as can be observed in the second scene. The information provided by the stack of SAR images does not seem sufficient to retrieve the ground height in this configuration. Including images with other incidence angles would help to estimate the heights in these locations.
}

To conduct a quantitative comparison of the segmentation results, we report the mean
error for each estimated surface to the ground truth \textit{cf.}
\ref{table:errors}. The TV parameter $\beta$ is set, for each
method, as the one minimizing this error.

\begin{table}[h]
	\begin{center}
		\begin{tabular}{ l  l  l  l }
			\hline
			\bf Estimator & \multicolumn{2}{c}{\bf Mean Error} &  $\boldsymbol\beta$ \\
			\cline{2-3}
			              & Scene a & Scene b &    \\
			\hline
			Capon Beamforming & 4.58 m & 5.84 m & 1.5 \\ 
			MUSIC & 3.23 m &  4.00 m  & 1.3 \\ 
			WSF & 3.12 m &  4.00 m  & 1.6 \\ 
			SPICE & 4.24 m &  4.21 m  & 12.6 \\ 
			3-D inversion & 2.50 m &  2.60 m  & 2.0 \\ 
			REDRESS & \textbf{1.60 m} &  \textbf{2.02 m}  & 2.0 \\
			\hline
		\end{tabular}	
	\end{center}
	\caption{Mean errors between the estimated surfaces
		and the ground truth, 
		last column: optimal $\beta$ values used for
		the surface segmentation.}
	\label{table:errors}
\end{table}

\section{Conclusion}

In this paper we introduced a graph-cut based segmentation algorithm to
estimate the urban surfaces from a SAR tomographic reconstruction. The proposed
approach is very general and can be used in combination with
many different tomographic algorithms. Experiments done on a set of 40 TerraSAR-X
images of Paris show good results for different tomographic estimators (Capon
beamforming, MUSIC, WSF, SPICE, CS and 3-D inversion).
As the 3-D inversion algorithm is designed to use 3-D priors, we also present
an algorithm that alternatively reconstructs the 3-D
distribution of reflectivities, segments the urban surfaces
from the volume of reflectivities and updates the
regularization so as to improve the subsequent 3-D
reconstruction. While the non-iterative 3-D inversion
algorithm fails in some cases to reduce the
main lobes of the strong scatterers, the alternating scheme
achieves a much sharper estimation of the distribution of
reflectivities.

\section{Acknowledgements}
This project has been funded by ANR (the French National
Research Agency) and DGA (Direction Générale de l’Armement) under ALYS
project ANR-15-ASTR-0002.

The TerraSAR-X images were provided by the DLR in the framework of the project LAN1746.

\newpage

\bibliographystyle{model2-names}
\bibliography{refs}

\end{document}